\documentclass[11pt]{article}

\usepackage[preprint]{acl}

\usepackage{times}
\usepackage{latexsym}
\usepackage[T1]{fontenc}
\usepackage[utf8]{inputenc}
\usepackage{microtype}
\usepackage{inconsolata}
\usepackage{graphicx}

\usepackage{amsmath}
\usepackage{amssymb}
\usepackage{booktabs}
\usepackage{multirow}
\usepackage{xspace}
\usepackage{xcolor}
\usepackage{colortbl}   
\usepackage{subcaption} 
\usepackage{algorithm}
\usepackage{algorithmic}
\usepackage{array}      

\definecolor{ourshl}{gray}{0.9}  

\graphicspath{{}{latex/}}

\newcommand{\method}{\textsc{Crafter}\xspace}
\newcommand{\wmape}{wMAPE\xspace}

\newcommand{\eg}{e.g.,\xspace}

\title{When Do Corrective Features Help? An Agent for Corrective Feature Discovery on Black-Box Forecasters}

\author{%
  \textbf{Fangxin Wang\textsuperscript{1}},
  \textbf{Ziyi Zhang\textsuperscript{2}},
  \textbf{Diyi Zhuang\textsuperscript{3}},
  \textbf{Langzhou He\textsuperscript{1}},
  \textbf{Shiyu Wang},
\\
  \textbf{Baichuan Mo\textsuperscript{4}},
  \textbf{Philip S. Yu\textsuperscript{1}}
\\
  \textsuperscript{1}University of Illinois Chicago,
  \textsuperscript{2}Texas A\&M University,
\\
  \textsuperscript{3}Massachusetts Institute of Technology,
  \textsuperscript{4}Tsinghua University
}

\hypersetup{
  pdftitle={When Do Corrective Features Help? An Agent for Corrective Feature Discovery on Black-Box Forecasters},
  pdfsubject={cs.LG},
  pdfkeywords={time series forecasting, automated feature engineering, LLM agent, residual correction, foundation models},
}

\begin{document}
\maketitle

\begin{abstract}
Frozen pretrained forecasters often fail in structured, recurring ways that are costly to repair through fine-tuning. We study \emph{corrective feature
discovery}: mining interpretable features of the frozen forecaster's residual to drive a lightweight
post-hoc corrector. Prior automated feature engineering models the data-generating process;
corrective features instead model the \emph{model-failure} process. We present \method{} (Corrective
Residual Agent with Feature-based Temporal Exploration and Reasoning), which keeps the
backbone frozen and mines its residual with two generators of different character: a
compositional search over the raw input channels, and a large language model (LLM) that proposes
named feature combinations, binary flags, and short executable code. A single validation-grounded gate
accepts or rejects every candidate blind to its origin, and a validation-selected corrector applies
the survivors or leaves the forecast unchanged. The same source-blindness lets prior
feature-engineering systems run through the identical pipeline, so \method{} doubles as an
instrument that attributes forecast changes to the feature source alone. Across six public datasets
and six frozen backbones, \method{} surpasses every dedicated feature-engineering system at every feature budget, roughly doubling the corrector-only lift and cutting the
weakest backbones' error by up to $27\%$. The gains are robust to the LLM backend and persist on top of fine-tuned backbones.
\end{abstract}

\section{Introduction}

Large-scale forecasting is now served by pretrained backbones deployed
\emph{frozen}~\citep{liu2024timer,ansari2024chronos,liu2025moirai}. An e-commerce platform or grocery
chain predicts sales for hundreds of thousands of store--product series from a single model, because
fine-tuning one per item or regime is infeasible. The frozen model still fails in structured,
recurring ways: it misses promotions, lags regime shifts, and underprices event spikes. These errors
are business-critical and must be interpretable to act on, so the question is not how to train a
better forecaster but how to \emph{correct} a deployed one cheaply, without modifying it.

A natural way to correct a fixed forecaster is to add external, interpretable features around it. This
connects the problem to automated feature engineering, but with a different target: once the backbone
is frozen, the object of discovery changes from the target process itself to the residual of a
particular deployed model. The relevant question is not which regularities help predict the future,
but which remain unmodeled by this backbone. We therefore study \emph{corrective feature discovery}:
automatically finding interpretable features of the residual, the signal of where and how a particular
frozen forecaster fails. The residual is not leftover noise a priori---it carries structure induced by
the backbone's blind spots, concentrated around promotions, rare events, and local regimes---so a
corrective feature is model-dependent and doubles as a diagnosis: each accepted feature names a
mechanism the deployed model misses.

Two properties make the discovery hard. First, the most valuable corrective features are \emph{named}
from domain semantics rather than enumerated from the raw schema; where they matter, a statistical
feature bank and a syntactic search improve the forecast only modestly, while named features close
most of the remaining gap (\S\ref{sec:exp-main}). Second, a feature's value is \emph{regime-dependent}:
it concentrates where the backbone errs and vanishes where the backbone is already accurate, so the
same source helps on one backbone--dataset cell and is inert or harmful on another. Whether a feature
source or corrector configuration helps therefore depends on the regime, and this dependence has not
been characterized systematically.

\method{} addresses the method and the characterization together. It keeps
the backbone frozen and
mines the residual with two generators of different character: a compositional search over the raw
channels, and an LLM that proposes named combinations, binary flags, and short executable code. A
single validation-grounded gate admits candidates blind to their origin, and a validation-selected
corrector applies the survivors or leaves the forecast unchanged when none help. This source-blind
design merges two heterogeneous generators into one method, and also turns the
pipeline into an instrument for fairly evaluating external feature-engineering system.

We evaluate \method{} on six public datasets and six frozen backbone families using a rolling-origin,
multi-seed protocol. All methods are run in the same evaluation harness and, where applicable, with
the same LLM backend. Across this controlled comparison, \method{} consistently outperforms three
dedicated feature-engineering systems. This advantage is robust to feature budget, remains under a
second LLM backend, and also holds when the corrected backbone is fine-tuned rather than used
zero-shot. On the two datasets where we test temporal reuse, the discovered features transfer across
backtest windows. Finally, we identify the limits of residual correction: in saturated cells, no
feature source yields a reliable improvement.

\paragraph{Contributions.}
\textbf{(C1) A source-blind framework for corrective feature discovery.}
We formulate feature discovery for a frozen forecaster as residual correction, and instantiate it
with a single validation-grounded gate that admits heterogeneous sources blind to their origin. The
framework is both a deployable correction layer and an evaluation instrument: any external
feature-engineering system runs through the identical gate, corrector, and budget, so a difference in
the forecast is attributable to the feature source alone.
\textbf{(C2) A characterization of when corrective features help.}
Across six datasets and six backbone families we map where residual correction is beneficial, inert,
or harmful. Correction pays in proportion to the residual's exploitable structure: large on weak
backbones with informative forecast covariates, where LLM-named features are the differentiator, and
neutral once a strong backbone saturates the series. Budget amplifies that margin rather than
creating it.


\section{Related Work}
\label{sec:related}

\paragraph{Frozen foundation forecasters.}
Pretrained time-series models are often deployed zero-shot, with a single backbone served frozen
across many series~\citep{ansari2024chronos,liu2024timer,liu2025moirai}. This avoids per-series
fine-tuning, but the frozen model leaves a backbone-specific residual it cannot remove on its own.
We take that residual as the object to model and leave the
backbone untouched, consistent with broader lightweight post-training approaches that add
capabilities to frozen foundation models without updating their pretrained
weights \citep{houlsby2019parameterefficienttransferlearningnlp, hu2021loralowrankadaptationlarge,
he2025molecularfoundationmodelsknow}.

\paragraph{Automated and LLM-based feature engineering.}
A long line of work builds features to improve a trainable predictor. Statistical banks extract many
candidate features and select among them~\citep{christ2018tsfresh,cerqueira2020vest,costa2021autofits},
and search-based engines compose operators over the raw schema with reinforcement
learning~\citep{khurana2018rlfe}, Monte-Carlo tree search~\citep{huang2022mctsfe}, or expert-level
generation~\citep{zhang2022openfe}. A recent line prompts an LLM to write feature code from dataset
context~\citep{hollmann2023caafe,nam2024octree,abhyankar2025llmfe,murray2025elate}, and AutoCT pairs
an LLM with tree search for tabular features~\citep{liu2025autoct}. Two things separate this work from
ours. Its objective is \emph{predictive}: the features train a downstream model rather than explain a
fixed one. And its strongest LLM methods target \emph{tabular} data. We run three of these systems as
feature generators inside our own pipeline (\S\ref{sec:exp-setup}), which puts every source on one
footing.

\paragraph{Correcting a frozen forecaster.}
Modeling the error of a fixed forecaster with a second learner is a classic
idea~\citep{zhang2003hybrid}. Recent methods correct a deployed forecaster without retraining it.
Post-Training Corrections picks a sequence of corrections from a fixed
pool~\citep{cherkaoui2025posttraining}. The $\delta$-Adapter learns a sparse input mask and a bounded
residual head, and uses no LLM~\citep{liang2026deltadapter}. Other methods adapt a frozen forecaster
online or through a memory module~\citep{lyu2026tsmemory,dai2026orca}, and AutoGluon-TimeSeries
regresses on raw covariates inside its pipeline~\citep{shchur2023autogluon}. These methods correct the
forecast, but they act on inputs that are already given. None \emph{discovers} new interpretable
features of the residual.

\paragraph{Discovering residual features to correct.}
Closest to our setting is NSR-Boost, which keeps a model frozen and uses an LLM to write symbolic
residual experts for an aggregator to combine~\citep{dai2026nsrboost}. It is tabular, it uses the LLM
as its only generator, and it runs no search. Work on forecasting explanation and interpretable
correction analyzes or adjusts model behavior, but does not discover corrective
features~\citep{wang2023forecastcf,lopez2024surrogate}. \method{} differs in three ways. It mines the
residual of a frozen \emph{forecaster}. It draws candidates from two different generators, a
compositional search and an LLM, and admits them through one validation gate that is blind to their
source. Because the gate is source-blind, the same pipeline also measures \emph{which} source and
configuration help in \emph{which} regime, which the work above does not study.


\begin{figure*}[t]
\centering
\includegraphics[width=\textwidth]          {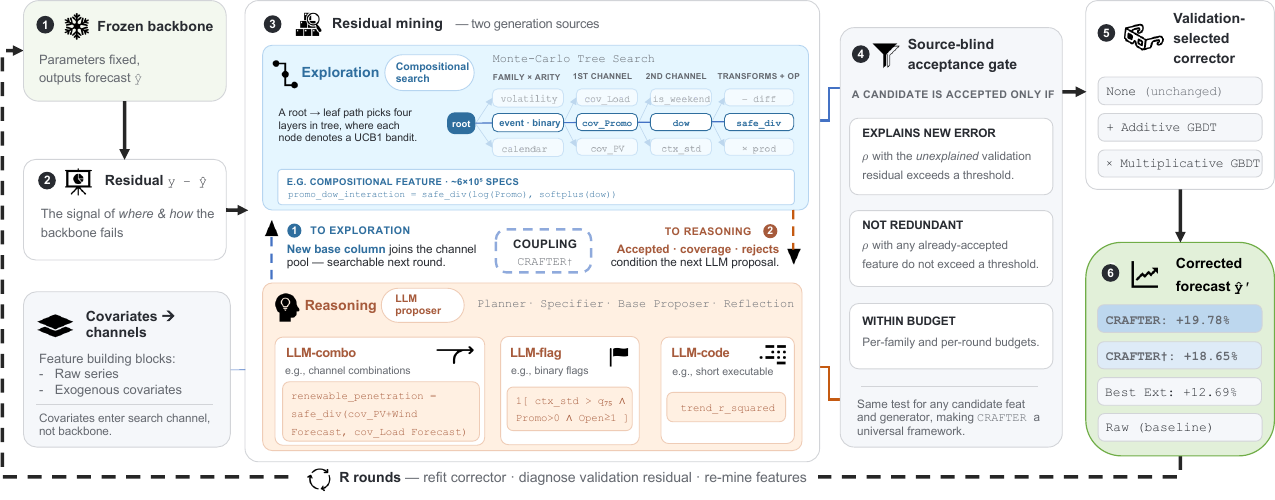}
  \caption{\textbf{The \method{} framework.} The forecaster stays frozen. Two generators mine its
  residual for structured feature specifications: a compositional Monte-Carlo tree search and an LLM
  proposer, optionally coupled (\method$^{\dagger}$). A source-blind gate admits a candidate only when
  it explains validation error the corrector cannot. A validation-selected corrector based on correlation $\rho$ and budget then applies the
  survivors, or leaves the forecast unchanged. The loop runs $R$ rounds.}
  \label{fig:crafter}
\end{figure*}

\section{Method}
\label{sec:method}

We present \method{} (\textbf{C}orrective \textbf{R}esidual \textbf{A}gent with \textbf{F}eature-based \textbf{T}emporal \textbf{E}xploration and \textbf{R}easoning), an agent for \emph{corrective feature
discovery}. It leaves the forecasting backbone frozen and mines interpretable features of its
residual, the signal of \emph{where} and \emph{how} the backbone fails, to drive a lightweight
post-hoc corrector. This section describes the framework and the single source-blind gate at its
core (\S\ref{sec:m-overview}), its two feature generators, \emph{Exploration} (a compositional search)
and \emph{Reasoning} (a language model), with their optional coupling (\S\ref{sec:m-generators}), the
gate (\S\ref{sec:m-gate}), and the validation-selected corrector (\S\ref{sec:m-zoo}).

\subsection{Overview}
\label{sec:m-overview}
Illustrated in Fig.~\ref{fig:crafter}, \method{} corrects a frozen forecaster without modifying it. Given the
backbone forecast $\hat y$
over a horizon, it mines interpretable features of the residual and fits a lightweight
gradient-boosted-tree corrector on them, which then adjusts the forecast or leaves it unchanged.
Quality is measured by the weighted mean absolute percentage error (\wmape{}) of the horizon total
(the corrector target and hyperparameters are in App.~A and~D).
Covariates enter only as corrector features, never as backbone inputs; \S\ref{sec:exp-robust} shows
this routing is what makes the gain possible.

Every candidate feature is a \emph{structured specification}: a typed record drawn from a fixed
grammar (App.~C) that applies transforms and an operator to a few input channels and
compiles to one numeric column. Specifications are searchable, de-duplicable across rounds, and
interpretable by name, so a run can be audited feature by feature. \method{} mines the residual with
two generators of different character---an atomic search over the raw channels and an LLM that
proposes named features (\S\ref{sec:m-llm})---and routes both into a single \emph{source-blind}
acceptance gate (\S\ref{sec:m-gate}) that keeps a candidate only when it explains validation error
the corrector cannot already account for, whichever generator produced it.

\method{} runs for $R$ rounds (Alg.~S1, App.~B). Each round fits the
corrector, diagnoses where it still errs on a held-out validation block, mines new candidates from
both generators, and keeps those that pass the gate. A validation-selected corrector then applies the
survivors or leaves the backbone unchanged when none help (\S\ref{sec:m-zoo}).

\subsection{Feature generators}
\label{sec:m-generators}
\method{} mines the residual with two generators of different character. A compositional search
composes the raw channels into new features, and an LLM names mechanisms the raw schema does not
contain. Both feed candidates into the single gate of \S\ref{sec:m-gate}, which decides what survives.

\paragraph{Compositional search.}
\label{sec:m-mcts}
The first generator builds \emph{compositional} features in the tradition of operator-based
automated feature engineering~\citep{cerqueira2020vest,costa2021autofits,zhang2022openfe}, named \textbf{atomic} in this paper. A feature
applies a transform to one channel, or an operator to two transformed channels, drawn from a fixed
vocabulary of $T{=}15$ transforms and $B{=}8$ operators (App.~C). The depth-one and
depth-two expressions number far too many to score exhaustively, about $6{\times}10^5$ for
$C{\approx}25$ channels, so \method{} searches. It uses a Monte-Carlo tree
search~\citep{kocsis2006uct} whose nodes are UCB1 bandits~\citep{auer2002ucb}. Channels are grouped
into mechanism \emph{families} such as \texttt{volatility} or \texttt{trend}; a family pools its
channels, so one reward updates the whole mechanism and the search finds productive families from few
samples before refining to a single channel. A root-to-leaf path picks a family and arity, then the
channels, then the operator, each node conditioned on the choices above it (App.~C). An
arm is credited by two signals: a cheap proxy at scoring time,
the rank correlation $\rho$ between the candidate and the still-unexplained validation residual, and the
realized gain in validation \wmape{} after the corrector is refit. The bandit selects arms by UCB1 and
updates each arm's value by the incremental sample mean of these rewards; the selection, update, and
reward rules are Eq.~(S3)--(S5) in App.~D. The realized gain ties the search
to the deployed model: a feature is worth what keeping it actually buys.

\paragraph{LLM-proposed features.}
\label{sec:m-llm} 
The search only recombines existing channels; the LLM \emph{invents} features that name mechanisms
the schema lacks. Organized into roles and queried a few times per round, it reasons over the run's own
evidence, not generic knowledge alone: which features the gate accepted, which it rejected, and which
mechanism families are still uncovered. A \textsc{Planner} reads this state and proposes a few
exploration \emph{directions}. A \textsc{Specifier} turns each direction into candidate specs built
from existing channels, surfacing semantic compositions a syntactic search would not prioritize: on
the \textsc{epf} electricity-price series it composes a \texttt{renewable\_penetration},
renewable generation over total load, which captures the price suppression a covariate-blind backbone
misses. A \textsc{Base Proposer} goes further and invents new named base
columns for mechanisms the schema lacks, which join the channel pool and become searchable next round.
A \textsc{Reflection} call revises the next round's directions when few candidates survive the gate.
The LLM proposes three kinds of features: 1) \textbf{LLM-combo}: combinations of existing channels, 2) \textbf{LLM-flag}: binary flags, and 3) \textbf{LLM-code}: short
executable code (accepted examples of every kind in App.~K). LLM only
\emph{proposes}: which candidates survive is decided by the gate of \S\ref{sec:m-gate}, not by the
model. All role prompts are in App.~H.

\paragraph{Coupling the two generators.}
\label{sec:m-bidir}
The two can also run as a coupled loop. A new base column invented by the LLM, once it passes the gate,
joins the channel pool, so the search can compose it with other channels in later rounds. Conversely,
each round's merged accepted set, family coverage, and rejected candidates condition the next LLM
proposal, steering it toward thin families and away from past rejects. Decided by whether allowing coupling, we have both \method and \method$^{\dagger}$ (coupled) evaluated in the experiments.

\subsection{The source-blind acceptance gate}
\label{sec:m-gate}
Both generators feed one gate, and that gate is what makes \method{} a single method rather than two
pipelines. A candidate is accepted only if it explains validation error the corrector cannot
\emph{already} account for: the Spearman correlation between the candidate and the unexplained
validation residual must exceed a threshold $\tau$, the candidate must not be collinear with an
already-accepted feature, and per-family and per-round budgets must hold. The same test applies to
every candidate, compositional or semantic. Deciding acceptance on one validation-grounded scale makes
the two generators commensurable without either reading the other's internal scores: the merged pool
self-selects against the residual, not against any generator's private confidence.

The gate is agnostic to a feature's origin, with a consequence beyond merging the two generators.
Any external feature-engineering system runs through the identical pipeline by swapping only the
generator, with the gate, corrector, and selection unchanged. \method{} then doubles as an
\emph{instrument}: a difference in the final forecast is attributable to the feature source, because
nothing else moves (\S\ref{sec:exp-main}). The same source-blindness that unifies the two generators
keeps that comparison fair.

\subsection{Validation-selected corrector}
\label{sec:m-zoo}

The features that pass the gate feed a corrector. \method{} fits a small \emph{corrector set}---an
identity (\textsc{None}) map plus an additive and a multiplicative gradient-boosted corrector---and
keeps the one with the best validation \wmape{}. Trees are chosen because the residual is nonlinear
and regime-dependent and because per-feature importances keep the correction auditable
(App.~D). The \textsc{None} option floors the risk \emph{on
validation} (a round never ships a corrector worse there than leaving the backbone alone), making
correction an asymmetric bet. This is a validation-time safeguard, not a test guarantee: under
rolling-origin shift a low-headroom cell can still regress out-of-sample (\eg \textsc{bizitobs}/Chronos,
\S\ref{sec:exp-main}).


\begin{figure*}[t]
\centering
\begin{subfigure}[t]{0.68\textwidth}
    \centering
    \includegraphics[width=\linewidth]{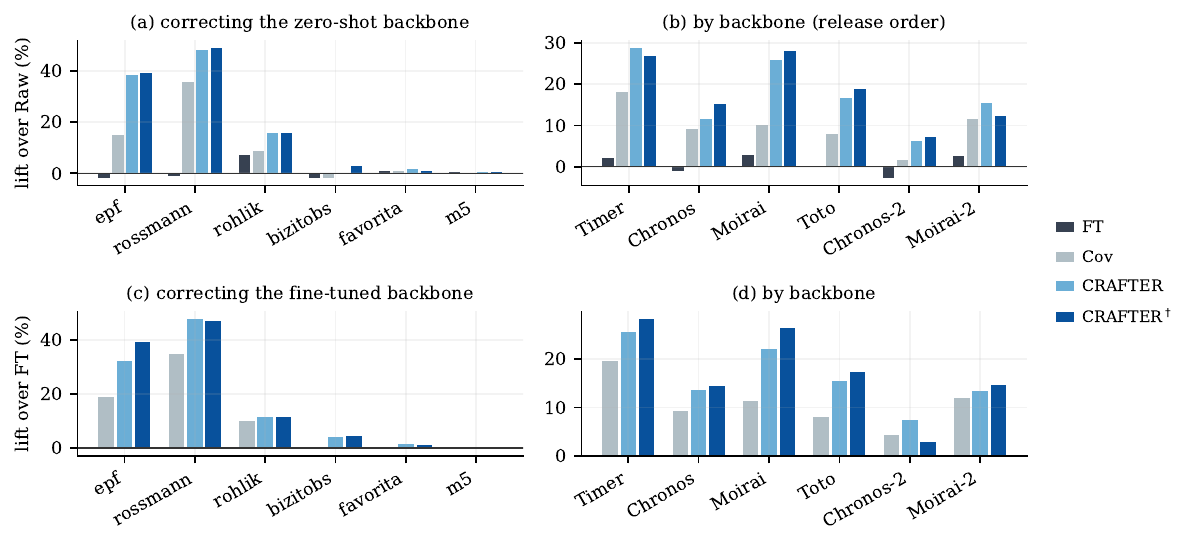}
    \caption{Correction before and after fine-tuning.}
    \label{fig:main}
\end{subfigure}
\hfill
\begin{subfigure}[t]{0.30\textwidth}
    \centering
    \includegraphics[width=\linewidth]{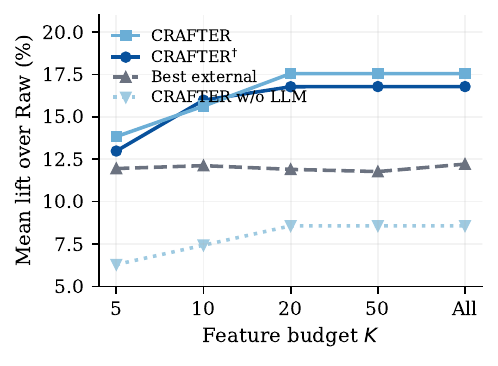}
    \caption{Same-budget fairness.}
    \label{fig:ksweep}
\end{subfigure}

\caption{\textbf{Correction holds across training and feature budgets.} Mean lift in test \wmape{}
(\%, higher is better) over the $6$ datasets. Colours are shared across panels; the two \method{}
variants share a hue. \textbf{(a)} Correcting the zero-shot backbone (lift over \textsc{Raw}) and the
fine-tuned one (lift over \textsc{FT}). \method{} helps in both, with the largest gains on weak
backbones. \textbf{(b)} Lift over \textsc{Raw} against budget $K$; ``All'' is uncapped. Both \method{}
variants beat the best external at every $K$; the no-LLM corrector stays below. Per-cell results in
App.~Table~S12.}
\label{fig:ft_ksweep}
\end{figure*}

\section{Experimental Setup}
\label{sec:exp-setup}

\paragraph{Datasets.}
We use six public forecasting datasets spanning the regimes where covariate information varies
in kind and strength (listed in the order used throughout): \textsc{epf-de} (day-ahead electricity
prices; a single long series with \emph{forecast-type} renewable covariates)~\citep{lago2021epf}, two
promotion-driven retail panels (\textsc{rossmann},
\textsc{rohlik})~\citep{kaggle2015rossmann,kaggle2024rohlik}, an hourly business-operations panel
(\textsc{bizitobs-l2c})~\citep{palaskar2024automixer}, a store-sales panel
(\textsc{favorita})~\citep{kaggle2017favorita}, and the M5 weekly competition data
(\textsc{m5})~\citep{makridakis2022m5}. Together they cover single- vs.\
many-series, covariate-rich vs.\ covariate-poor, and weak- vs.\ saturated-backbone settings.

\paragraph{Frozen backbones.}

Our grid corrects six pretrained backbones used zero-shot, listed in order of release:
Timer~\citep{liu2024timer}, Chronos~\citep{ansari2024chronos}, Moirai~\citep{woo2024moirai}, and Toto~\citep{cohen2024toto}, followed by Chronos-2~\citep{ansari2025chronos2} and
Moirai-2.0~\citep{liu2025moirai}, spanning the first foundation-model wave to current leaders. Chronos and Timer are primarily univariate or channel-independent forecasters; Moirai and Toto can handle multivariate time-series inputs; and Chronos-2 explicitly exposes a covariate-aware forecasting interface. In every grid we nonetheless run
\emph{all} backbones covariate-blind, so the backbone forecast never sees the covariates and every
method corrects the \emph{same} residual---any covariate value is then attributable to the corrector,
not the backbone (routing them into the backbone instead is at best neutral; \S\ref{sec:exp-robust}). The
grid is $6\times6=36$ cells.

\paragraph{Protocol.}
We use a rolling-origin backtest: on cached forecasts we re-cut three expanding train:val:test splits
($6{:}1{:}1$, $7{:}1{:}1$, $8{:}1{:}1$) with no test leakage, validate on three blocked CV windows, run
$R{=}3$ rounds, and report last-round test \wmape{} averaged over the three splits and three seeds
(nine runs per cell). We assess significance with a one-sided Wilcoxon signed-rank test over the paired
cells and decompose run-to-run variance into seed and split components (App.~D); split
variance dominates. The public datasets may appear in either LLM's pretraining corpus, but the leakage
surface is small: the LLM emits only feature \emph{specifications} and code reading past history and
the backbone forecast, never test-period values. Because
seeds do not change the cached forecasts, the grid is GPU-free and cheap; compute and timing details are
in App.~D.

\paragraph{Baselines and backend.}
Every feature-engineering method---both \method{} variants and the external systems---uses the same
GPT-5.2 backend and an identical harness (same cached residual, same corrector set, same validation
top-$K$ selection), so only the feature \emph{generator} differs and the comparison is fair by
construction. We report two \method{} variants, the two rightmost (highlighted) columns of
App.~Table~S12: \method{}, the full system, and \method$^{\dagger}$, which additionally enables
the two source-coupling channels of \S\ref{sec:m-bidir}. The baselines are: the frozen backbone
(\textsc{Raw}); a backbone fine-tune (\textsc{FT}; a uniform head-only fine-tune of every backbone, with a
stronger full-parameter Moirai variant in App.~F); a covariate-only corrector
(\textsc{Cov}) and the no-LLM corrector (\method-noLLM), which ablate the LLM and the search; and three
external feature-engineering systems adapted to our setting---\textsc{TSFresh} (a statistical feature
bank)~\citep{christ2018tsfresh}, CAAFE (LLM-written feature code)~\citep{hollmann2023caafe}, and LLM-FE
(LLM-guided evolutionary search)~\citep{abhyankar2025llmfe}. App.~E states what each
adaptation keeps and changes.

\section{Results}

\label{sec:exp-results}
Corrective feature discovery improves a frozen forecaster without retraining it, and the size of that
improvement is governed by residual headroom rather than by backbone identity, feature budget, or LLM
backend. We evaluate on six datasets and six pretrained backbones under one rolling-origin,
multi-seed protocol in which every feature-engineering method passes the same validation gate and
downstream corrector, so measured differences trace to the discovered features rather than to the
corrector (App.~D; full per-cell grid in Table~S12). This section
establishes five points: \method{} is state of the art against prior feature engineering and
helps weak backbones most (\S\ref{sec:exp-main}); the wrapper generalizes across backbone families
(\S\ref{sec:exp-when}) and survives non-standard input paths (\S\ref{sec:exp-robust}); the decisive
gain comes from the LLM (\S\ref{sec:exp-why}); and the learned features transfer across time
(\S\ref{sec:exp-transfer}).

\begin{figure*}[t]
\centering

\begin{minipage}[t]{\dimexpr\textwidth-\columnwidth-1em\relax}
\vspace{0pt}
\centering
\captionof{table}{\textbf{Cross-LLM robustness.} Per-backbone summary under both LLM backends. Both
variants improve every backbone, most on the weak ones, so the backbone-strength picture is
LLM-agnostic; dataset-level increments still depend on the backend (\S\ref{sec:exp-why}).
\textsc{Raw} is the backbone's mean zero-shot \wmape{}$\times100$ over the $6$ datasets.
\textbf{Lift\%} is the mean reduction vs.\ \textsc{Raw}; \textbf{wins} counts datasets (of $6$) where
the variant matches or beats the best external.}
\label{tab:xllm}

\footnotesize
\setlength{\tabcolsep}{5pt}
\resizebox{\linewidth}{!}{%
\begin{tabular}{lc cc cc cc cc}
\toprule
 & & \multicolumn{4}{c}{GPT-5.2} & \multicolumn{4}{c}{DeepSeek-3.4-pro} \\
\cmidrule(lr){3-6}\cmidrule(lr){7-10}
 & & \multicolumn{2}{c}{\method} & \multicolumn{2}{c}{\method$^{\dagger}$} & \multicolumn{2}{c}{\method} & \multicolumn{2}{c}{\method$^{\dagger}$} \\
\cmidrule(lr){3-4}\cmidrule(lr){5-6}\cmidrule(lr){7-8}\cmidrule(lr){9-10}
Backbone & \textsc{Raw} & Lift\% & wins & Lift\% & wins & Lift\% & wins & Lift\% & wins \\
\midrule
Timer    & $42.7$ & $+25.7$ & 5/6 & $+25.8$ & 5/6 & $+23.6$ & 3/6 & $+29.7$ & 4/6 \\
Chronos & $31.7$ & $+8.5$  & 4/6 & $+11.4$ & 5/6 & $+9.6$  & 4/6 & $+9.5$  & 3/6 \\
Moirai & $43.9$ & $+27.2$ & 5/6 & $+26.8$ & 5/6 & $+11.5$ & 2/6 & $+27.9$ & 5/6 \\
Toto     & $35.4$ & $+14.8$ & 4/6 & $+16.6$ & 4/6 & $+11.0$ & 4/6 & $+15.0$ & 6/6 \\
Chronos-2 & $31.3$ & $+6.4$ & 4/6 & $+7.4$ & 4/6 & $+5.8$ & 3/6 & $+11.9$ & 5/6 \\
Moirai-2 & $34.0$ & $+12.5$ & 3/6 & $+12.8$ & 3/6 & $+12.3$ & 4/6 & $+16.7$ & 5/6 \\
\bottomrule
\end{tabular}%
}
\end{minipage}
\hfill
\begin{minipage}[t]{\columnwidth}
\vspace{0pt}
\centering
\includegraphics[width=\columnwidth]{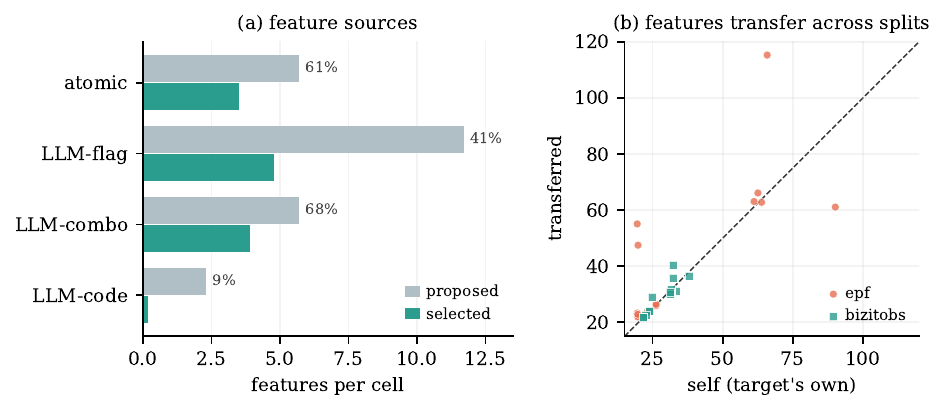}
\captionof{figure}{\textbf{Feature provenance and transfer.} \textbf{(a)} Features proposed vs.\
selected by source, with survival rate. LLM kinds supply about three-quarters of the selected set,
and LLM-code is almost always pruned. \textbf{(b)} Features learned on one split and replayed on a
later one track the target split's own features along the diagonal. The off-diagonal points are the
Moirai shift corner, mostly single-series \textsc{epf} (\S\ref{sec:exp-transfer};
App.~L).}
\label{fig:analysis}
\end{minipage}

\end{figure*}

\subsection{Corrective discovery improves frozen and fine-tuned backbones}
\label{sec:exp-main}

Figure~\ref{fig:main} summarizes the main comparison across six datasets and six backbone families;
the full per-cell grid with external baselines is in App.~Table~S12. Under the shared
harness of \S\ref{sec:exp-setup}, both \method{} variants outperform each of the three dedicated
feature-engineering systems, with per-baseline paired Wilcoxon tests significant at $p{<}0.01$.
The result holds both when aggregating by dataset and when aggregating by backbone (Figure \ref{fig:main} and Table~\ref{tab:xllm}), showing that corrective discovery is not a single-dataset artifact.

\textbf{The gain is not a feature-budget artifact.}
The same-budget sweep rules the explanation that \method{} wins simply because it is allowed to propose more candidates. In Fig.~\ref{fig:ksweep} (per-budget win rates and mean differences in App.~Table~S11),
\method{} beats the best external feature-engineering system at every feature budget $K$. Even at $K{=}5$, \method{} exceeds the best
external system at its own uncapped budget, and the margin saturates by moderate budgets rather than
appearing only in the ``All'' setting. The external systems' curve is comparatively flat, indicating
that their additional candidates do not translate into comparable residual-correction gains under the
shared gate and corrector. Thus the main result reflects feature quality, not merely feature count.

\textbf{Fine-tuning is not a substitute for residual correction.}
We also compare against a uniform head-only fine-tune, \textsc{FT}, applied identically across
backbones. Fine-tuning improves the frozen forecasts, but it does not remove the structured residuals
that \method{} exploits. Correcting the zero-shot backbone beats fine-tuning alone on most cells, and
applying \method{} on top of the fine-tuned backbone still yields additional lift over \textsc{FT}
(Fig.~\ref{fig:main}, top vs.\ bottom; the full per-cell zero-shot-versus-fine-tuned grid is
App.~Table~S1). The same conclusion holds for the stronger full-parameter
in-domain fine-tune of the covariate-capable Moirai models, reported in App.~Table~S2.
These results place \method{} as a deployment-time correction layer rather than a replacement for
backbone adaptation: it improves a frozen model, and it remains useful after fine-tuning when
exploitable residual structure remains.

\subsection{Search matters through coupling, not only standalone lift}
\label{sec:exp-why}

We next discuss where the gain in \S\ref{sec:exp-main} comes from. The source-blind gate in the pipeline serve as an attribution device: external feature-engineering systems, compositional search,
and LLM-generated candidates are all filtered by the same validation rule and passed to the same
corrector. 

\textbf{Search defines the enumerable feature space.}
The compositional search adds little standalone lift on these public benchmarks (the noLLM column of
Table~S12; a component-level with/without-LLM comparison is App.~Table~S5),
but this null result is informative rather than incidental. It shows what can be recovered by syntactic enumeration over
the observed schema. When this search and the external feature banks fail to close the residual gap,
but named LLM features do, the missing signal is not merely another lag, rolling aggregate, or
low-order interaction. It is a semantic feature that must be named in a form aligned with the
backbone's residual. The search arm therefore acts as a control for the LLM comparison: it measures
the enumerable floor, so the remaining gain is identified as a semantic feature gap rather than
assumed to be one.

The clearest case is \textsc{epf}. External banks and syntactic search add little beyond the
covariate-only corrector, while \method{} removes much more of the frozen error using named
forecast-covariate features such as \texttt{renewable\_penetration} (the \textsc{epf} rows of
Table~S12; representative accepted features in App.~K). This is the regime
corrective discovery is designed for: the relevant information is present in the covariates, but it
does not become useful until it is expressed as a feature that matches the backbone's failure mode.
Similar, though smaller, effects appear on multi-series retail panels, where domain-named
interactions generalize across related store--product series.

\textbf{Coupling turns search from a control into a mechanism.}
The coupled variant, \method$^{\dagger}$, enables the two cross-feeding generators in
\S\ref{sec:m-bidir}: search results summarize which enumerable candidates survived the gate, and that
feedback is exposed to the LLM in later proposal rounds. Search alone adds little net lift, yet
feeding its accepted and rejected candidates back to the LLM makes \method$^{\dagger}$ stronger on
balance: it improves on \method{} on more cells, gives per-backbone lift at least as large on all but
one backbone, and has its clearest gains on \textsc{epf} (per-backbone lift in Table~\ref{tab:xllm};
per-cell in Table~S12). Since the search contributes little alone, that gain is the
cross-feeding steering the LLM toward features it would not otherwise propose. Its cost is variance,
especially on volatile single-series cells such as \textsc{epf}, where a small change in the feature
set can move a cell in either direction (seed versus split variance in App.~D). We
therefore report \method{} and \method$^{\dagger}$ as two endpoints of an explore--exploit trade-off.

\textbf{Feature-level evidence supports the same mechanism.}
The provenance analysis in Fig.~\ref{fig:analysis}a shows that most selected features come from
LLM-generated semantic sources, while atomic search features survive at a comparable rate but are
limited by the smaller enumerable pool (per-source and per-dataset counts in App.~J). This matches the performance attribution: the LLM supplies
most of the selected feature mass, while search marks the reliable enumerable candidates and provides
feedback for coupling. Code-generated features are rarely selected, probably constrained by limited program synthesis. Instead, \method{} works by combining three roles: a corrector that
tests whether a feature actually reduces residual error, a search process that establishes and feeds
back the enumerable space, and an LLM that names semantic features outside that space.

\subsection{When do corrective features help?}
\label{sec:exp-when}

The central condition is residual headroom, and App.~Fig.~S1 is its most direct support:
plotting every backbone--dataset cell's corrected gain against the backbone's own \textsc{Raw} error,
\method's lift concentrates where the raw error is large and collapses to zero where the backbone has
already saturated the series. Corrective features help when the frozen backbone leaves structured
errors that external signals can explain; they are neutral or harmful otherwise. This condition is
stronger than backbone identity, release order, or feature budget.

\textbf{Headroom, not backbone identity, determines the gain.}
The clearest evidence comes from comparing the same dataset across backbone families. On
\textsc{bizitobs}, correction gives large lift when the raw backbone is weak, as with Timer and
Moirai, but becomes neutral or negative when the raw forecast already saturates the series, as
with Chronos and Moirai-2.0. The gain is governed by how much residual structure the backbone leaves
behind, not by which foundation model produced the forecast. This also explains why newer or stronger backbones can still
benefit on some datasets, such as \textsc{rossmann} and \textsc{epf}, while older or weaker backbones
do not automatically improve everywhere. 

\textbf{The useful feature source depends on the kind of residual structure.}
When the remaining error is tied to semantic forecast covariates, named features matter most. This is
the case on \textsc{epf}, where external feature banks and syntactic search add little beyond the
covariate-only corrector, but LLM-named features close much more of the frozen error (the \textsc{epf}
rows of Table~S12, whose selected set is LLM-dominated in App.~Table~S7).
When the available covariates are already directly informative, as on the retail panels, the
covariate-only corrector already captures a large share of the gain and additional searched features
are often redundant (the \textsc{rossmann} and \textsc{favorita} rows of Table~S12). When
the residual contains no stable structure, no feature source reliably helps and every column collapses
onto \textsc{Raw} (the saturated \textsc{m5} rows, and the flat right-hand region of
App.~Fig.~S1). The question is therefore not whether one source is globally best, but which
source matches the residual left by a particular backbone on a particular dataset.

\textbf{Budget amplifies headroom but does not create it.}
The same-budget sweep gives the feature-budget axis of the characterization (Fig.~\ref{fig:ksweep};
App.~Table~S11). Increasing $K$ widens
\method's margin in cells where residual headroom exists, and the gains saturate by moderate budgets.
But a larger budget does not turn saturated cells into wins: if the backbone has already absorbed the
available structure, more candidate features mostly add variance and selection risk. Budget is
therefore an amplifier, not a switch. It helps exploit residual structure when that
structure is present, but cannot manufacture structure where the residual is already close to
noise.
The practical guidance is therefore simple: deploy corrective
feature discovery when validation residuals remain structured, especially in weak-backbone or
semantic-covariate regimes; treat feature budget as a way to scale gains in those regimes; and avoid
forcing correction in saturated cells where no feature source passes the validation gate reliably.

\subsection{Robustness to backend and input path}
\label{sec:exp-robust}

The preceding sections use a fixed LLM backend and route all auxiliary information through the
corrector. We check that the method and the headroom characterization survive two deployment choices:
which LLM proposes features, and where covariates or multivariate inputs are injected.

\textbf{Changing the LLM backend preserves the characterization.}
Repeating the full grid with DeepSeek-3.4-pro yields the same qualitative pattern as the main
backend: both \method{} variants improve every backbone on average, and the gains remain largest where
the raw backbone leaves more residual headroom (Table~\ref{tab:xllm}; per-cell results in
App.~Table~S13). Thus the conclusion of \S\ref{sec:exp-when} is not an artifact of one
LLM. The backend does affect individual cells, especially volatile single-series cases such as
\textsc{epf}, so we do not claim backend-invariant feature sets or identical per-dataset increments.
The robust claim is weaker and more useful: changing the LLM changes which named features are found,
but it does not change the deployment rule that correction helps when exploitable residual structure
remains.

\textbf{Covariate-capable backbones do not replace the corrector.}
A second deployment choice is whether to feed covariates directly to a backbone that accepts them, or
to keep the backbone forecast-only and route covariates to the residual corrector. The direct route
does not dominate. Toto ingests covariates harmlessly, with little change in \textsc{Raw}, while
Moirai-2.0 can degrade substantially when given the same inputs directly
(App.~Table~S3). In contrast, the correction layer is stable across this choice:
on the retail panels, \method{} reaches similar corrected accuracy whether or not the covariates were
first exposed to the backbone, and it can recover even when covariate ingestion degrades the raw
forecast. This supports the design choice in \S\ref{sec:m-overview}: keep the backbone fixed and route
auxiliary signals to a learnable residual corrector.

\subsection{Learned features transfer across time}
\label{sec:exp-transfer}

A corrective feature is useful only if it captures persistent structure rather than a single-window
artifact. We test this by \emph{replaying} the features accepted on one split onto a later split, with
the LLM and search turned off, so only the feature set's origin changes; the target split's own
features serve as an upper bound, and a covariate-only corrector as a lower bound. Across two
datasets, five backbones, and every ordered split pair, transferred features match the target's own
features at the median and stay within a few percent on most cells (Fig.~\ref{fig:analysis}b;
App.~Table~S9). Performance decays only mildly with split distance, while
catastrophic failures concentrate in the Moirai shift corner, especially the volatile single-series
\textsc{epf}. Despite low overlap in accepted feature names across splits, performance transfers
because different runs often discover functionally equivalent features. Corrective feature discovery
therefore recovers reusable residual structure, not per-window noise.



\section{Conclusion}
\label{sec:conclusion}

We studied corrective feature discovery: correcting a frozen black-box forecaster by mining
interpretable features of its residual instead of retraining it. \method{} mines the residual with a
compositional search and a language model, and admits their candidates through one validation-grounded
gate that is blind to a feature's origin. The shared gate turns two generators into a single method
and lets any external feature-engineering system run through the identical pipeline, so \method{} also
serves as an instrument that attributes a change in the forecast to the feature source alone.
Correction pays in proportion to the residual's exploitable structure: large on weak backbones with
informative forecast covariates, where LLM-named features are the differentiator, and neutral once a
strong backbone already fits the series---we report the saturated cells alongside the wins. The
finding holds across six backbones and two LLM backends, is not bought with a larger feature budget,
and survives fine-tuning. Because features learned on one window transfer to the next on the datasets
we test (\S\ref{sec:exp-transfer}), a natural next step is a lightweight feature-memory that carries
mined features forward to amortize the search and warm-start the corrector, adapting as a backbone's
errors evolve across windows.

\bibliography{custom} 

@article{cerqueira2020vest,
  title={VEST: Automatic Feature Engineering for Forecasting},
  author={Cerqueira, Vitor and Moniz, Nuno and Soares, Carlos},
  journal={arXiv preprint arXiv:2010.07137},
  year={2020}
}

@mastersthesis{costa2021autofits,
  title={Autofits: Automated Feature Engineering for Irregular Time-Series},
  author={Costa, Pedro Miguel Pinto},
  school={Universidade do Porto},
  year={2021}
}

@article{hollmann2023caafe,
  title={Large Language Models for Automated Data Science: Introducing CAAFE for Context-Aware Automated Feature Engineering},
  author={Hollmann, Noah and M{\"u}ller, Samuel and Hutter, Frank},
  journal={arXiv preprint arXiv:2305.03403},
  year={2023}
}

@article{nam2024octree,
  title={Optimized Feature Generation for Tabular Data via LLMs with Decision Tree Reasoning},
  author={Nam, Jaehyun and Kim, Kyuyoung and Oh, Seunghyuk and Tack, Jihoon and Kim, Jaehyung and Shin, Jinwoo},
  journal={arXiv preprint arXiv:2406.08527},
  year={2024}
}

@article{abhyankar2025llmfe,
  title={LLM-FE: Automated Feature Engineering for Tabular Data with LLMs as Evolutionary Optimizers},
  author={Abhyankar, Nikhil and Shojaee, Parshin and Reddy, Chandan K.},
  journal={arXiv preprint arXiv:2503.14434},
  year={2025}
}

@article{wang2023forecastcf,
  title={Counterfactual Explanations for Time Series Forecasting},
  author={Wang, Zhendong and Miliou, Ioanna and Samsten, Isak and Papapetrou, Panagiotis},
  journal={arXiv preprint arXiv:2310.08137},
  year={2023}
}

@article{lopez2024surrogate,
  title={Surrogate Modeling for Explainable Predictive Time Series Corrections},
  author={Lopez, Alfredo and Sobieczky, Florian},
  journal={arXiv preprint arXiv:2412.19897},
  year={2024}
}

@article{murray2025elate,
  title={ELATE: Evolutionary Language model for Automated Time-series Engineering},
  author={Murray, Andrew and Dervovic, Danial and Cashmore, Michael},
  journal={arXiv preprint arXiv:2508.14667},
  year={2025}
}

@inproceedings{liu2024timer,
  title={Timer: generative pre-trained transformers are large time series models},
  author={Liu, Yong and Zhang, Haoran and Li, Chenyu and Huang, Xiangdong and Wang, Jianmin and Long, Mingsheng},
  booktitle={Proceedings of the 41st International Conference on Machine Learning},
  pages={32369--32399},
  year={2024}
}

@article{
ansari2024chronos,
title={Chronos: Learning the Language of Time Series},
author={Abdul Fatir Ansari and Lorenzo Stella and Ali Caner Turkmen and Xiyuan Zhang and Pedro Mercado and Huibin Shen and Oleksandr Shchur and Syama Sundar Rangapuram and Sebastian Pineda Arango and Shubham Kapoor and Jasper Zschiegner and Danielle C. Maddix and Hao Wang and Michael W. Mahoney and Kari Torkkola and Andrew Gordon Wilson and Michael Bohlke-Schneider and Bernie Wang},
journal={Transactions on Machine Learning Research},
issn={2835-8856},
year={2024},
url={https://openreview.net/forum?id=gerNCVqqtR},
note={Expert Certification}
}

@misc{ansari2025chronos2,
title={Chronos-2: From Univariate to Universal Forecasting},
author={Abdul Fatir Ansari and Oleksandr Shchur and Jaris K{\"u}ken and Andreas Auer and Boran Han and Pedro Mercado and Syama Sundar Rangapuram and Huibin Shen and Lorenzo Stella and Xiyuan Zhang and Mononito Goswami and Shubham Kapoor and Danielle C. Maddix and Pablo Guerron and Tony Hu and Junming Yin and Nick Erickson and Prateek Mutalik Desai and Hao Wang and Huzefa Rangwala and George Karypis and Yuyang Wang and Michael Bohlke-Schneider},
year={2025},
eprint={2510.15821},
archivePrefix={arXiv},
primaryClass={cs.LG},
doi={10.48550/arXiv.2510.15821},
url={https://arxiv.org/abs/2510.15821}
}

@article{liu2025moirai,
  title={Moirai 2.0: When less is more for time series forecasting},
  author={Liu, Chenghao and Aksu, Taha and Liu, Juncheng and Liu, Xu and Yan, Hanshu and Pham, Quang and Savarese, Silvio and Sahoo, Doyen and Xiong, Caiming and Li, Junnan},
  journal={arXiv preprint arXiv:2511.11698},
  year={2025}
}

@article{auer2002ucb,
  title={Finite-time Analysis of the Multiarmed Bandit Problem},
  author={Auer, Peter and Cesa-Bianchi, Nicol{\`o} and Fischer, Paul},
  journal={Machine Learning},
  volume={47},
  number={2--3},
  pages={235--256},
  year={2002},
  publisher={Springer}
}

@inproceedings{kocsis2006uct,
  title={Bandit Based Monte-Carlo Planning},
  author={Kocsis, Levente and Szepesv{\'a}ri, Csaba},
  booktitle={Machine Learning: ECML 2006},
  pages={282--293},
  year={2006},
  publisher={Springer}
}

@article{zhang2022openfe,
  title={{OpenFE}: Automated Feature Generation with Expert-level Performance},
  author={Zhang, Tianping and Zhang, Zheyu and Fan, Zhiyuan and Luo, Haoyan and Liu, Fengyuan and Liu, Qian and Cao, Wei and Li, Jian},
  journal={arXiv preprint arXiv:2211.12507},
  year={2022}
}

@article{christ2018tsfresh,
  title={Time Series {FeatuRe} Extraction on basis of Scalable Hypothesis tests (tsfresh -- A {Python} package)},
  author={Christ, Maximilian and Braun, Nils and Neuffer, Julius and Kempa-Liehr, Andreas W.},
  journal={Neurocomputing},
  volume={307},
  pages={72--77},
  year={2018},
  publisher={Elsevier}
}

@inproceedings{khurana2018rlfe,
  title={Feature Engineering for Predictive Modeling using Reinforcement Learning},
  author={Khurana, Udayan and Samulowitz, Horst and Turaga, Deepak},
  booktitle={Proceedings of the AAAI Conference on Artificial Intelligence},
  volume={32},
  number={1},
  year={2018}
}

@inproceedings{huang2022mctsfe,
  title={Automatic Feature Engineering Through {Monte Carlo} Tree Search},
  author={Huang, Yiran and Zhou, Yexu and Hefenbrock, Michael and Riedel, Till and Fang, Likun and Beigl, Michael},
  booktitle={Machine Learning and Knowledge Discovery in Databases (ECML PKDD)},
  series={Lecture Notes in Computer Science},
  volume={13715},
  year={2022},
  publisher={Springer}
}

@article{liu2025autoct,
  title={{AutoCT}: Automating Interpretable Clinical Trial Prediction with {LLM} Agents},
  author={Liu, Fengze and Wang, Haoyu and Cho, Joonhyuk and Roth, Dan and Lo, Andrew W.},
  journal={arXiv preprint arXiv:2506.04293},
  year={2025}
}

@article{zhang2003hybrid,
  title={Time series forecasting using a hybrid {ARIMA} and neural network model},
  author={Zhang, G. Peter},
  journal={Neurocomputing},
  volume={50},
  pages={159--175},
  year={2003},
  publisher={Elsevier}
}

@inproceedings{shchur2023autogluon,
  title={{AutoGluon-TimeSeries}: {AutoML} for Probabilistic Time Series Forecasting},
  author={Shchur, Oleksandr and Turkmen, Ali Caner and Erickson, Nick and Shen, Huibin and Shirkov, Alexander and Hu, Tony and Wang, Bernie},
  booktitle={Proceedings of the Second International Conference on Automated Machine Learning (AutoML)},
  year={2023}
}

@article{cherkaoui2025posttraining,
  title={Post-Training Corrections for Improved Time-Series Forecasting},
  author={Hamza Cherkaoui and Malik Tiomoko and Giuseppe Paolo and Yili Zhang and Meng Yu and Keli Zhang and Hafiz {Tiomoko Ali}},
  journal={arXiv preprint arXiv:2505.15354},
  year={2025}
}

@article{liang2026deltadapter,
  title={The Forecast After the Forecast: A Post-Processing Shift in Time Series},
  author={Daojun Liang and Qi Li and Yinglong Wang and Jing Chen and Hu Zhang and Xiaoxiao Cui and Qizheng Wang and Shuo Li},
  journal={arXiv preprint arXiv:2601.20280},
  year={2026}
}

@article{lyu2026tsmemory,
  title={{TS-Memory}: Plug-and-Play Memory for Time Series Foundation Models},
  author={Sisuo Lyu and Siru Zhong and Tiegang Chen and Weilin Ruan and Qingxiang Liu and Taiqiang Lv and Qingsong Wen and Raymond Chi-Wing Wong and Yuxuan Liang},
  journal={arXiv preprint arXiv:2602.11550},
  year={2026}
}

@article{dai2026orca,
  title={Learning the Context of Errors: Black-Box Online Adaptation of Time Series Foundation Models},
  author={Xilin Dai and Yiding Liu and Hongjie Xia and Yifan Hu and Zewei Dong and Jiang-Ming Yang and Qiang Xu},
  journal={arXiv preprint arXiv:2606.14222},
  year={2026}
}

@article{dai2026nsrboost,
  title={{NSR-Boost}: A Neuro-Symbolic Residual Boosting Framework for Industrial Legacy Models},
  author={Ziming Dai and Dabiao Ma and Jinle Tong and Mengyuan Han and Jian Yang and Hongtao Liu and Haojun Fei and Qing Yang},
  journal={arXiv preprint arXiv:2601.10457},
  year={2026}
}

@inproceedings{woo2024moirai,
  title={Unified Training of Universal Time Series Forecasting Transformers},
  author={Woo, Gerald and Liu, Chenghao and Kumar, Akshat and Xiong, Caiming and Savarese, Silvio and Sahoo, Doyen},
  booktitle={Proceedings of the 41st International Conference on Machine Learning (ICML)},
  year={2024}
}

@article{cohen2024toto,
  title={Toto: Time Series Optimized Transformer for Observability},
  author={Cohen, Ben and Khwaja, Emaad and Wang, Kan and Masson, Charles and Ram{\'e}, Elise and Doubli, Youssef and Abou-Amal, Othmane},
  journal={arXiv preprint arXiv:2407.07874},
  year={2024}
}

@article{lago2021epf,
  title   = {Forecasting day-ahead electricity prices: A review of state-of-the-art algorithms, best practices and an open-access benchmark},
  author  = {Lago, Jesus and Marcjasz, Grzegorz and De Schutter, Bart and Weron, Rafa{\l}},
  journal = {Applied Energy},
  volume  = {293},
  pages   = {116983},
  year    = {2021},
  doi     = {10.1016/j.apenergy.2021.116983}
}

@article{makridakis2022m5,
  title   = {M5 accuracy competition: Results, findings, and conclusions},
  author  = {Makridakis, Spyros and Spiliotis, Evangelos and Assimakopoulos, Vassilios},
  journal = {International Journal of Forecasting},
  volume  = {38},
  number  = {4},
  pages   = {1346--1364},
  year    = {2022},
  doi     = {10.1016/j.ijforecast.2021.11.013}
}

@article{palaskar2024automixer,
  title   = {AutoMixer for Improved Multivariate Time-Series Forecasting on Business and IT Observability Data},
  author  = {Palaskar, Santosh and Ekambaram, Vijay and Jati, Arindam and Gantayat, Neelamadhav and Saha, Avirup and Nagar, Seema and Nguyen, Nam H. and Dayama, Pankaj and Sindhgatta, Renuka and Mohapatra, Prateeti and Kumar, Harshit and Kalagnanam, Jayant and Hemachandra, Nandyala and Rangaraj, Narayan},
  journal = {Proceedings of the AAAI Conference on Artificial Intelligence},
  volume  = {38},
  number  = {21},
  pages   = {22962--22968},
  year    = {2024},
  doi     = {10.1609/aaai.v38i21.30336}
}

@misc{kaggle2015rossmann,
  author       = {{Kaggle}},
  title        = {Rossmann Store Sales},
  year         = {2015},
  howpublished = {Kaggle competition, \url{https://www.kaggle.com/c/rossmann-store-sales}}
}

@misc{kaggle2017favorita,
  author       = {{Kaggle}},
  title        = {Corporaci{\'o}n Favorita Grocery Sales Forecasting},
  year         = {2017},
  howpublished = {Kaggle competition, \url{https://www.kaggle.com/c/favorita-grocery-sales-forecasting}}
}

@misc{kaggle2024rohlik,
  author       = {{Kaggle}},
  title        = {Rohlik Sales Forecasting Challenge},
  year         = {2024},
  howpublished = {Kaggle competition, \url{https://www.kaggle.com/competitions/rohlik-sales-forecasting-challenge}}
}

@misc{he2025molecularfoundationmodelsknow,
      title={Can Molecular Foundation Models Know What They Don't Know? A Simple Remedy with Preference Optimization}, 
      author={Langzhou He and Junyou Zhu and Fangxin Wang and Junhua Liu and Haoyan Xu and Yue Zhao and Philip S. Yu and Qitian Wu},
      year={2025},
      eprint={2509.25509},
      archivePrefix={arXiv},
      primaryClass={cs.LG},
      url={https://arxiv.org/abs/2509.25509}, 
}

@misc{hu2021loralowrankadaptationlarge,
      title={LoRA: Low-Rank Adaptation of Large Language Models}, 
      author={Edward J. Hu and Yelong Shen and Phillip Wallis and Zeyuan Allen-Zhu and Yuanzhi Li and Shean Wang and Lu Wang and Weizhu Chen},
      year={2021},
      eprint={2106.09685},
      archivePrefix={arXiv},
      primaryClass={cs.CL},
      url={https://arxiv.org/abs/2106.09685}, 
}

@misc{houlsby2019parameterefficienttransferlearningnlp,
      title={Parameter-Efficient Transfer Learning for NLP}, 
      author={Neil Houlsby and Andrei Giurgiu and Stanislaw Jastrzebski and Bruna Morrone and Quentin de Laroussilhe and Andrea Gesmundo and Mona Attariyan and Sylvain Gelly},
      year={2019},
      eprint={1902.00751},
      archivePrefix={arXiv},
      primaryClass={cs.LG},
      url={https://arxiv.org/abs/1902.00751}, 
}

%
\clearpage
\appendix

\setcounter{table}{0}\renewcommand{\thetable}{S\arabic{table}}
\setcounter{figure}{0}\renewcommand{\thefigure}{S\arabic{figure}}
\setcounter{algorithm}{0}\renewcommand{\thealgorithm}{S\arabic{algorithm}}
\setcounter{equation}{0}\renewcommand{\theequation}{S\arabic{equation}}

\section*{Appendix}
\noindent Tables, figures, algorithms and equations below carry an S prefix
(Table~S1, Fig.~S1, Alg.~S1, Eq.~(S1)); plain references (\S3.2, Fig.~2a,
Table~1) point to the main paper above.


\section{Correction Target and Metric}
\label{app:task}
For each series and forecast origin we observe a history $x_{1:T}$, a frozen backbone forecast
$\hat{y}_{1:H}$ over horizon $H$, and (at training time) the truth $y_{1:H}$. \method{} predicts a
per-series multiplicative correction factor
\begin{equation}
\label{eq:factor}
\begin{split}
\tilde{y}_{1:H}&=f\cdot\hat{y}_{1:H},\\
f&=\mathrm{clip}\!\Big(\tfrac{\sum_h y_h}{\sum_h \hat{y}_h},\,\ell,\,u\Big),
\quad [\ell,u]=[0.3,3.0],
\end{split}
\end{equation}
with the clip admitting genuine multiplicative structure (\eg promotions, price changes) while
rejecting heavy-tailed ratios from near-zero denominators; an additive-offset target
$f^{+}=\sum_h y_h-\sum_h\hat{y}_h$ is also supported and selected against within the corrector set
(\S3.4). We report the weighted mean absolute percentage error of the horizon total,
\begin{equation}
\label{eq:wmape}
\wmape=\frac{\sum_i\big|\sum_h \tilde{y}_{i,h}-\sum_h y_{i,h}\big|}
            {\sum_i\big|\sum_h y_{i,h}\big|}.
\end{equation}
We score the horizon \emph{total} because it is the quantity deployment consumes---replenishment,
procurement, and capacity decisions are driven by aggregate demand over the lead time---and the
quantity the correction factor of Eq.~\eqref{eq:factor} acts on. The comparison stays fair: every
method, the frozen backbone included, is scored on the same total.

\section{The \method{} Round}
\label{app:round}
Alg.~\ref{alg:round} gives one round of the loop sketched in \S3.1. A run repeats it
for $R$ rounds, threading the accepted-feature set $\mathcal{F}$, the search arm statistics $\Theta$,
and the channel pool $\mathcal{C}$ across rounds. The two generators of \S3.2
appear as \emph{Exploration} (compositional UCB search, \S3.2) and \emph{Reasoning}
(the LLM roles, \S3.2); both feed the single source-blind gate of \S3.3
(line~\ref{alg:line-gate}), which keeps only candidates whose Spearman correlation with the
\emph{unexplained} validation residual exceeds $\tau$, that are not collinear with an accepted
feature, and that fit the per-family and per-round budgets. After refitting, a per-tier permutation
step prunes the lowest-value features by $\Delta$validation-\wmape{} (line~\ref{alg:line-trim}), and
the kept/trimmed outcome plus the realized validation gain is credited back to the search arms as
UCB feedback. The cross-round coupling shown---LLM-named base columns entering $\mathcal{C}$
so the search can compose them (line~\ref{alg:line-couple})---is one of the two channels
\method$^{\dagger}$ enables (\S3.2); it is off in the decoupled \method{}.

\begin{algorithm}[b]
\caption{One \method{} round}
\label{alg:round}
\begin{algorithmic}[1]
\REQUIRE accepted features $\mathcal{F}$, arm statistics $\Theta$, channel pool $\mathcal{C}$, frozen backbone forecast $\hat{y}$
\STATE fit the corrector set on $\mathcal{F}$; keep the head (\textsc{None}/additive/multiplicative) with lowest validation \wmape{} \COMMENT{\textsc{None} floors risk, \S3.4}
\STATE $r \leftarrow$ unexplained validation residual under the kept corrector; compute family coverage
\STATE $D \leftarrow \textsc{Planner}(\text{coverage},\,\mathcal{F},\,\text{blacklist})$ \COMMENT{Reasoning: exploration directions, \S3.2}
\STATE $P_{\mathrm{R}} \leftarrow \big(\textstyle\bigcup_{d\in D}\textsc{Specifier}(d)\big)\cup\textsc{BaseProposer}()$ \COMMENT{Reasoning: LLM-named features}
\STATE $P_{\mathrm{E}} \leftarrow \textsc{UCB-Sample}(\Theta,\,\mathcal{C})$ \COMMENT{Exploration: compositional search, \S3.2}
\STATE $\mathcal{A} \leftarrow \big\{\,c\in\mathrm{dedup}(P_{\mathrm{E}}\cup P_{\mathrm{R}}) : |\rho(c,r)|>\tau,\ \text{non-collinear},\ \text{within budgets}\,\big\}$ \label{alg:line-gate} \COMMENT{source-blind gate, \S3.3}
\STATE $\mathcal{C} \leftarrow \mathcal{C}\cup\{\text{new base columns in }\mathcal{A}\}$ \label{alg:line-couple} \COMMENT{Reasoning$\to$Exploration coupling (off by default)}
\STATE refit the corrector on $\mathcal{F}\cup\mathcal{A}$; $\mathcal{T} \leftarrow$ features below the per-tier top-$K$ by permutation $\Delta$val-\wmape{} \label{alg:line-trim} \COMMENT{prune}
\STATE update $\Theta$ with UCB credit from kept vs.\ trimmed importances and the validation-\wmape{} gain
\RETURN $(\mathcal{F}\cup\mathcal{A})\setminus\mathcal{T},\ \Theta,\ \mathcal{C}$
\end{algorithmic}
\end{algorithm}

\section{Feature Grammar}
\label{app:grammar}
A corrective feature is a structured specification: a typed record that compiles deterministically
to one numeric column over items. Specifications take one of three types, and every field is drawn
from a fixed vocabulary, so the space is bounded, searchable, and canonicalizable. The atomic search
emits \texttt{combo} records; the LLM emits \texttt{combo}, \texttt{flag}, and \texttt{code} records,
which the rest of the paper names \textbf{LLM-combo}, \textbf{LLM-flag}, and \textbf{LLM-code}.

\begin{itemize}\itemsep1pt
\item \textbf{\texttt{combo}} --- a pairwise expression
  $\textsc{post}\big(\tau_a(c_a)\,\oplus\,\tau_b(c_b)\big)$, where $c_a,c_b$ are input channels
  (base features or covariates), $\tau_\bullet$ are unary transforms, $\oplus$
  is a binary operator, and \textsc{post} is an optional output transform with an optional clip
  (a single transformed channel when $c_b,\oplus$ are omitted).
\item \textbf{\texttt{flag}} --- a $0/1$ indicator formed by combining a small set of thresholded
  conditions $\{(c,\,\mathrm{op},\,v)\}$ with \texttt{all}/\texttt{any}.
\item \textbf{\texttt{code}} --- a short Python function over the target history and the backbone
  forecast, restricted by an abstract-syntax-tree allow-list to \texttt{numpy}, \texttt{math}, and
  \texttt{scipy.\{signal,stats,fft,special\}} with no I/O, \texttt{exec}, or \texttt{eval}.
\end{itemize}

\noindent\textbf{Vocabulary.} The search ranges over $B{=}8$ binary operators
$\oplus\in\{\texttt{add},\texttt{sub},\texttt{mul},\texttt{div},\texttt{abs\_diff},\texttt{min},
\texttt{max},\texttt{safe\_div}\}$ and $T{=}15$ unary transforms $\tau \in$ \{\texttt{null},\texttt{neg},
\texttt{abs},\texttt{log},\texttt{log1p},\texttt{signlog},\texttt{sqrt},\texttt{square},\texttt{exp},
\texttt{sigmoid},\texttt{tanh},\texttt{softplus},\texttt{zscore},\texttt{winsorize},\texttt{ema} \};
flag conditions draw from the operators $\{<,\le,>,\ge,=,\neq,\texttt{between},\texttt{in},
\texttt{not\_in}\}$. The executor additionally admits a few rarer transforms and operators
(\eg \texttt{geo\_mean}, quantile summaries); these are reachable by the LLM but outside the
atomic search's UCB vocabulary.

\noindent\textbf{Feature families.} Each channel belongs to one mechanism \emph{family}, the unit
L1 routes over (\S3.2). Representative families and example channels:
\texttt{level} (\texttt{mean\_sales}, \texttt{zero\_ratio}); \texttt{trend} (\texttt{trend\_ratio},
\texttt{trend\_consistency}); \texttt{seasonality} (\texttt{autocorr7}, \texttt{seasonal\_strength});
\texttt{volatility} (\texttt{cv}, \texttt{outlier\_rate}); \texttt{shift} (\texttt{regime\_shift},
\texttt{shift\_magnitude}); \texttt{anomaly} (\texttt{spike\_ratio}, \texttt{consecutive\_zeros}).

\noindent\textbf{Example.} The JSON record \texttt{\{type:combo, feat\_a:cv, feat\_b:mean\_sales,
op:div, transform\_a:log, transform\_b:null\}} denotes $\log(\mathrm{cv})/\mathrm{mean\_sales}$ and
is named accordingly. A \emph{canonical key} maps equivalent records to one identifier: the display
name and all bookkeeping fields (source, family, scores) are dropped, the empty/\texttt{null}/identity
transforms collapse to a single sentinel, numeric thresholds and clips are rounded, and \texttt{flag}
conditions are sorted, so the key is an order- and rename-invariant hash of the remaining grammar
fields. It drives cross-round de-duplication and a rejection blacklist that suppresses any candidate
already tried and refused.

\section{Corrector and Hyperparameters}
\label{app:hparams}
The corrector is a histogram gradient-boosted regressor ($200$ trees, max depth $4$, learning rate
$0.1$, min-samples-leaf $20$). The corrector set comprises $\{\textsc{None},\text{additive},
\text{multiplicative-factor}\}$ correctors, the best selected by validation \wmape{}; the
multiplicative-factor head is the strongest on weak-backbone $\times$ forecast-covariate cells, where a
per-row learned factor is more flexible than per-horizon offsets. We fix its clip to $[0.3,3.0]$ rather
than a dataset-adaptive bound: an adaptive bound is occasionally tighter on clean retail but blows up on
weak-backbone $\times$ LLM cells, so we trade a little adaptivity to remove that failure mode.
Acceptance gate: a candidate is kept iff $|\rho_{\mathrm{Spearman}}(\text{feature},\,
\text{unexplained residual})|\ge\tau$, pairwise $|\rho|\le 0.75$, with at most two features per
family and ten per round. The threshold is a single global constant $\tau{=}0.05$, fixed across all
datasets rather than tuned per series, so there is no per-dataset grid. Search defaults: a
four-layer conditioned UCB1 tree with exploration constant $\kappa{=}1.4$, $R{=}3$ rounds, and
$K{=}4$ planner directions; the selection, update, and reward rules follow.

\paragraph{Bandit selection and update.}
The compositional search is a Monte-Carlo tree of UCB1 bandits with one arm per choice at each
layer (family, then arity, then channels, then operator). Every arm $a$ stores a visit count
$n_a$ and a running mean reward $\hat{\mu}_a$; each L1 family arm keeps one $(n_a,\hat{\mu}_a)$
pair per arity, so univariate and bivariate evidence never mix. At a node whose children are
$\mathcal{A}$ with total visits $N=\sum_{a'\in\mathcal{A}} n_{a'}$, the descent selects the arm
that maximizes the UCB1 score,
\begin{equation}
a^{\star}=\arg\max_{a\in\mathcal{A}}\ \hat{\mu}_a+\kappa\sqrt{\frac{\ln N}{n_a}},
\label{eq:ucb}
\end{equation}
visiting any never-tried arm ($n_a{=}0$) first. Once an arm's candidate has been scored, or a
round has finished, the arm is credited with a reward $r$ and updated by the incremental sample
mean
\begin{equation}
n_a \leftarrow n_a+1,
\qquad
\hat{\mu}_a \leftarrow \hat{\mu}_a+\frac{1}{n_a}\bigl(r-\hat{\mu}_a\bigr),
\label{eq:banditupdate}
\end{equation}
applied to every arm on the selected root-to-leaf path; a decayed copy is also propagated to
sibling contexts that share the family or base channel. The reward combines two signals. A fast
per-candidate proxy is available already at scoring time,
\begin{equation}
r_{\mathrm{proxy}}=\alpha\,\lvert\rho\rvert+\beta\,\mathbb{I}[\text{accepted}],
\qquad \alpha{=}0.4,\ \beta{=}0.3,
\label{eq:banditreward}
\end{equation}
where $\rho$ is the Spearman correlation between the candidate and the still-unexplained
validation residual and $\mathbb{I}[\text{accepted}]$ marks passing the acceptance gate above.
After the corrector is refit at each round's end, a realized signal supplements the proxy: every
surviving feature is credited in proportion to its gradient-boosting importance, plus a
$\gamma{=}0.3$ round-level term on the round's validation-\wmape{} improvement. The proxy steers
exploration cheaply within a round; the realized signal ties the search to the deployed corrector,
so a feature is ultimately worth what keeping it actually buys.

\paragraph{Why gradient-boosted trees.}
The residual is nonlinear and regime-dependent, concentrating in promotions, regime shifts, and rare
events, so a linear corrector cannot fit it. Gradient-boosted trees have the capacity and keep the
correction auditable: a per-feature importance names \emph{which} mechanism drives a backbone's
residual. Read across backbones, the corrector \emph{diagnoses} each model's blind spots, naming what
a given backbone systematically misses (\S5.2).

\paragraph{Protocol and compute.}
\label{app:protocol}
All backbone forecasts are computed once on NVIDIA A100/A40 GPUs and cached; because random seeds do not
change a frozen backbone's forecast, the entire multi-seed grid (seeds $42$--$44$ $\times$ three rolling
splits per cell) reuses the warm cache with \emph{no} GPU recomputation, and the residual-correction loop
itself runs on CPU. The stack is Python 3.12, with PyTorch 2.x for the backbone runners and LightGBM 4.x
for the corrector. A full single-seed sweep of the grid completes in ${\sim}30$ minutes on warm caches---orders of
magnitude cheaper than per-dataset backbone fine-tuning, which requires a GPU training run for every
(dataset, backbone) pair. We report last-round ($R{=}3$) test \wmape{}, averaged over the three splits
and three seeds, and decompose variance into seed (optimizer/LLM-sampling) and split (temporal-drift)
components; split variance dominates. Quantified per cell on the DeepSeek grid (\wmape{}$\times100$):
the median seed-std of \method{} is $0.49$ against a median split-std of ${\sim}3.2$; the frozen
backbone is seed-deterministic, the no-LLM and \textsc{TSFresh} columns sit near $0.02$, and the LLM
methods' seed noise concentrates on \textsc{epf}.

\paragraph{Per-arity statistics and the univariate transform.} Each L1 family arm stores two
independent (count, mean) pairs, one per arity, so univariate and bivariate evidence never mix.
We do not search the univariate transform: with only $T{=}15$ options and that branch serving mainly
to surface new channels for later composition, a bandit there would chiefly drain samples from the
bivariate branch, where the $B\,T^2$ operator-triple choice carries the reward. Because the bivariate
branch is sampled far more often, its visit count grows much faster; we periodically rebalance the
two arities' counts so the exploration bonus $\kappa\sqrt{\ln N/n}$ does not lock onto the
under-sampled univariate arity.

\section{External-System Adaptation}
\label{app:externals}
The three external systems run \emph{inside} the \method{} harness: the same cached backbone
residual, the same base and covariate columns, the same gradient-boosted corrector and corrector-set
selection, and the same validation top-$K$ feature selector. Only the candidate generator is
swapped, so the comparison isolates feature generation (\S4). Each generator is
kept as close to its published form as the forecasting setting allows.
\begin{itemize}\itemsep1pt
\item \textbf{TSFresh} extracts its standard statistical bank over each series history through the
  library's own extractor (${\sim}30$ statistics per series), from which the shared selector keeps
  the top $K$.
\item \textbf{CAAFE} targets classification, so its LLM iteration loop's validation-accuracy
  acceptance test runs on the residual ratio discretized into three bins (over-, well-, and
  under-predicted); the accepted features themselves remain continuous columns for the shared
  corrector.
\item \textbf{LLM-FE} runs its evolutionary loop, with LLM-proposed mutations and crossovers over a
  population of feature expressions and fitness the validation correlation against the same
  residual target.
\end{itemize}
The two LLM-based systems use the same backend, per-run seed, and feature budget as \method{}. We do
not run the systems' native end-to-end pipelines, because they target tabular prediction with their
own downstream model: a native run would change the corrector, the target, and the metric at once,
and a difference in the final forecast could then no longer be attributed to the feature source.
This inside-the-harness design is the point of the instrument (\S3.3); its cost is that
we measure each system as a feature generator for residual correction, not its native end-to-end
performance.

\section{Fine-tuning baseline}
\label{app:ft}
The \textsc{FT} baseline in Table~\ref{tab:master} applies a \emph{uniform head-only} fine-tune to every
backbone (training the output head with the backbone frozen), early-stopping on validation \wmape{} with
the test window untouched, on splits identical to \method{}; a constant recipe keeps the comparison fair
across backbones. As a stronger check, the two Moirai backbones, which ship an official recipe,
additionally receive a full-parameter in-domain fine-tune (Table~\ref{tab:ft}): it reuses the official
\texttt{uni2ts} \textsc{MoiraiFinetune} module and loss but masks the last $H$ patches (eval-aligned)
rather than at random, selects checkpoints on validation \wmape{} rather than NLL, and trains in fp32
with gradient clipping and warmup. The comprehensive grid (Table~\ref{tab:ftgrid}; all six backbones
$\times$ six datasets, both backbone branches $\times$ four correction gradients) reports this head-only
\textsc{FT} next to each correction level. The grids are single-seed (42), split-averaged, so absolute
values differ slightly from the $3$-seed main table, most visibly on the volatile single-series
\textsc{epf}; the before/after comparison of Fig.~2a (top vs.\ bottom) uses both branches
within the same seed and is unaffected. Fine-tuning requires a GPU training run for every (dataset, backbone) pair,
orders of magnitude more costly than the GPU-free correction loop (App.~\ref{app:protocol}).

\begin{table*}[htbp]
\centering \scriptsize \setlength{\tabcolsep}{4.5pt}
\caption{\textbf{Fine-tuning does not remove the gain (full grid).} Test \wmape{}$\times100$ (lower is
better) for the four correction gradients---no correction (\textsc{Raw} zero-shot / \textsc{FT}
fine-tuned), covariate-only corrector (\textsc{Cov}), \method{}, and \method$^{\dagger}$---on the
zero-shot and the fine-tuned backbone, over all $6$ backbones $\times$ $6$ datasets (\textsc{chr}$_2$ is
covariate-aware Chronos-2, single-seed; its covariate-only corrector was not re-run on the fine-tuned
branch, so the two \textsc{Cov} columns carry the same value). Fine-tuning is the
uniform head-only probe (full-parameter Moirai in Table~\ref{tab:ft}). On the non-saturated datasets
(\textsc{epf}, \textsc{rossmann}, \textsc{rohlik}) the ordering $\method{}<\textsc{Cov}<$ no-correction
holds on \emph{both} branches; \textsc{m5} is saturated and all gradients
coincide. The \method{} and \method$^{\dagger}$ columns (\colorbox{ourshl}{gray}) are highlighted in
both branches. \textbf{Bold} marks the best gradient within each branch (a \method{} variant when it
ties). Single seed (42), split-averaged.}
\label{tab:ftgrid}
\resizebox{0.85\textwidth}{!}{%
\begin{tabular}{ll cc>{\columncolor{ourshl}}c>{\columncolor{ourshl}}c cc>{\columncolor{ourshl}}c>{\columncolor{ourshl}}c}
\toprule
& & \multicolumn{4}{c}{Zero-shot backbone} & \multicolumn{4}{c}{Fine-tuned backbone (head-only)} \\
\cmidrule(lr){3-6}\cmidrule(lr){7-10}
Dataset & BB & \textsc{Raw} & \textsc{Cov} & \method & \method$^{\dagger}$ & \textsc{FT} & \textsc{Cov} & \method & \method$^{\dagger}$ \\
\midrule
\multirow{6}{*}{epf} & tim & 53.7 & 46.2 & \textbf{30.2} & \textbf{30.2} & 56.9 & 43.5 & \textbf{31.1} & \textbf{31.2} \\
 & chr & 49.9 & 45.4 & \textbf{32.0} & \textbf{31.7} & 52.0 & 44.2 & \textbf{32.6} & \textbf{30.2} \\
 & moi & 73.2 & 60.7 & \textbf{36.8} & \textbf{28.4} & 71.2 & 56.4 & \textbf{44.5} & \textbf{30.2} \\
 & toto & 55.8 & 49.8 & \textbf{31.6} & \textbf{28.8} & 55.8 & 49.8 & \textbf{40.4} & \textbf{30.7} \\
 & chr$_2$ & 52.6 & 43.7 & \textbf{37.1} & \textbf{32.2} & 55.1 & 43.7 & \textbf{42.4} & 55.0 \\
 & moi$_2$ & 55.0 & 41.6 & \textbf{38.6} & 50.7 & 55.0 & 41.8 & 42.1 & \textbf{29.7} \\
\midrule
\multirow{6}{*}{rossmann} & tim & 39.5 & 16.7 & \textbf{13.5} & \textbf{13.5} & 39.5 & 16.8 & \textbf{13.7} & \textbf{13.4} \\
 & chr & 20.6 & 11.9 & \textbf{11.3} & \textbf{11.3} & 21.3 & 14.0 & \textbf{13.4} & \textbf{13.2} \\
 & moi & 30.0 & 24.7 & \textbf{13.4} & \textbf{13.1} & 29.8 & 24.7 & \textbf{13.0} & \textbf{13.1} \\
 & toto & 22.2 & 14.5 & \textbf{12.5} & \textbf{12.2} & 22.2 & 14.4 & \textbf{12.2} & \textbf{12.2} \\
 & chr$_2$ & 20.4 & 15.9 & \textbf{13.1} & \textbf{12.8} & 21.1 & 15.9 & \textbf{12.5} & \textbf{14.0} \\
 & moi$_2$ & 21.4 & 12.6 & \textbf{11.7} & \textbf{11.7} & 21.4 & 12.6 & \textbf{12.1} & \textbf{11.7} \\
\midrule
\multirow{6}{*}{rohlik} & tim & 36.1 & 33.4 & \textbf{28.6} & \textbf{28.6} & 30.3 & 27.2 & \textbf{26.7} & \textbf{26.7} \\
 & chr & 27.4 & 24.7 & \textbf{24.4} & \textbf{24.4} & 27.2 & 24.5 & \textbf{24.3} & \textbf{24.3} \\
 & moi & 36.2 & 33.5 & \textbf{28.5} & \textbf{28.4} & 31.2 & 28.3 & \textbf{27.6} & \textbf{27.6} \\
 & toto & 28.2 & 25.1 & \textbf{24.8} & \textbf{24.9} & 28.2 & 25.3 & \textbf{25.1} & \textbf{25.2} \\
 & chr$_2$ & 26.6 & 23.8 & \textbf{23.6} & \textbf{23.6} & 27.1 & \textbf{23.8} & 24.0 & 23.9 \\
 & moi$_2$ & 34.9 & 32.2 & \textbf{28.0} & \textbf{28.0} & 29.2 & 26.3 & \textbf{25.4} & \textbf{25.4} \\
\midrule
\multirow{6}{*}{bizitobs} & tim & 49.5 & 42.4 & \textbf{36.7} & \textbf{41.9} & 51.9 & 41.8 & 42.3 & \textbf{34.3} \\
 & chr & 24.6 & 25.7 & 30.1 & \textbf{24.6} & \textbf{25.0} & 25.6 & 26.1 & 25.9 \\
 & moi & 51.5 & 43.5 & \textbf{39.1} & \textbf{39.3} & 50.9 & 42.3 & \textbf{40.2} & \textbf{36.8} \\
 & toto & 39.0 & 40.9 & \textbf{37.6} & \textbf{35.3} & 39.0 & 40.9 & \textbf{35.2} & \textbf{37.2} \\
 & chr$_2$ & \textbf{23.6} & 31.0 & 31.1 & 31.4 & \textbf{25.3} & 31.0 & 30.6 & 30.4 \\
 & moi$_2$ & 27.1 & 27.7 & 27.3 & \textbf{26.3} & 27.1 & 27.7 & \textbf{26.6} & 31.1 \\
\midrule
\multirow{6}{*}{favorita} & tim & 24.9 & 21.8 & \textbf{21.4} & \textbf{21.7} & 23.2 & 21.6 & \textbf{20.5} & \textbf{20.5} \\
 & chr & 17.5 & 17.6 & \textbf{17.5} & 17.6 & 17.4 & 17.6 & \textbf{17.1} & \textbf{17.4} \\
 & moi & 20.6 & 19.8 & \textbf{19.7} & \textbf{19.7} & 20.6 & \textbf{19.8} & 19.9 & 19.9 \\
 & toto & 17.4 & 17.7 & 17.5 & \textbf{17.4} & 17.4 & 17.7 & \textbf{17.4} & 17.5 \\
 & chr$_2$ & \textbf{14.4} & 15.2 & 15.3 & 15.9 & \textbf{14.6} & 15.2 & 15.7 & 15.4 \\
 & moi$_2$ & 16.5 & 16.6 & \textbf{16.5} & \textbf{16.5} & 16.5 & 16.6 & 16.6 & \textbf{16.5} \\
\midrule
\multirow{6}{*}{m5} & tim & 52.5 & \textbf{51.0} & 51.1 & 51.2 & 51.4 & \textbf{50.8} & 50.9 & 50.9 \\
 & chr & 50.3 & 50.3 & \textbf{50.1} & \textbf{50.0} & 49.9 & 49.9 & 50.0 & \textbf{49.9} \\
 & moi & 52.2 & 52.1 & \textbf{51.2} & \textbf{50.9} & 52.2 & 52.1 & \textbf{51.0} & \textbf{51.0} \\
 & toto & \textbf{49.7} & 50.1 & 49.9 & 50.0 & \textbf{49.7} & 50.1 & 49.9 & 50.0 \\
 & chr$_2$ & 50.4 & 51.2 & \textbf{50.2} & \textbf{50.3} & \textbf{49.5} & 51.2 & 50.3 & 50.3 \\
 & moi$_2$ & \textbf{49.0} & 49.3 & 49.5 & 49.5 & \textbf{49.0} & 49.3 & 49.3 & 49.5 \\
\bottomrule
\end{tabular}}
\end{table*}

\begin{table*}[htbp]
\centering
\caption{\textbf{Full-parameter Moirai stress test.} The four correction gradients on a \emph{fully}
fine-tuned Moirai backbone---no correction (\textsc{FT}), covariate-only corrector (\textsc{Cov}),
\method{}, and \method$^{\dagger}$---alongside the zero-shot branch, for the two covariate-capable Moirai
models (test \wmape{}$\times100$, lower is better). \method{} still helps on $6$ of $8$ fine-tuned cells;
fine-tuning rarely beats correcting the frozen backbone. The \method{}/\method$^{\dagger}$ columns
(\colorbox{ourshl}{gray}) are highlighted; \textbf{bold} marks a corrector that beats both no-correction
baselines in its branch (\textsc{Raw}/\textsc{Cov} on the left, \textsc{FT}/\textsc{Cov} on the right),
the Table~\ref{tab:master} rule. Exceptions: \textsc{favorita}, \textsc{bizitobs} under
Moirai-2.0. Single seed (42), split-averaged.}
\label{tab:ft}
\renewcommand{\arraystretch}{1.2}
\resizebox{0.85\textwidth}{!}{%
\begin{tabular}{ll cc>{\columncolor{ourshl}}c>{\columncolor{ourshl}}c cc>{\columncolor{ourshl}}c>{\columncolor{ourshl}}c}
\toprule
& & \multicolumn{4}{c}{Zero-shot backbone} & \multicolumn{4}{c}{Fine-tuned backbone (full-param)} \\
\cmidrule(lr){3-6}\cmidrule(lr){7-10}
Dataset & BB & \textsc{Raw} & \textsc{Cov} & \method & \method$^{\dagger}$ & \textsc{FT} & \textsc{Cov} & \method & \method$^{\dagger}$ \\
\midrule
\multirow{2}{*}{epf} & moi & 73.2 & 60.8 & \textbf{36.7} & \textbf{28.4} & 60.8 & 55.5 & \textbf{37.8} & \textbf{44.4} \\
 & moi$_2$ & 55.0 & 41.6 & \textbf{38.6} & 45.8 & 55.0 & 41.6 & \textbf{32.2} & \textbf{29.0} \\
\midrule
\multirow{2}{*}{rossmann} & moi & 30.0 & 26.2 & \textbf{13.2} & \textbf{13.0} & 26.1 & 24.9 & \textbf{17.2} & \textbf{16.4} \\
 & moi$_2$ & 21.4 & 12.6 & \textbf{11.8} & \textbf{12.0} & 21.4 & 12.6 & \textbf{11.9} & \textbf{11.6} \\
\midrule
\multirow{2}{*}{bizitobs} & moi & 51.5 & 43.5 & \textbf{38.5} & \textbf{39.5} & 55.1 & 53.5 & \textbf{40.7} & \textbf{38.8} \\
 & moi$_2$ & 27.1 & 30.1 & 31.0 & 30.5 & 27.1 & 30.1 & 29.6 & 28.9 \\
\midrule
\multirow{2}{*}{favorita} & moi & 20.6 & 20.3 & \textbf{20.1} & 20.4 & 25.4 & 23.3 & \textbf{22.5} & \textbf{22.4} \\
 & moi$_2$ & 16.5 & 16.6 & 16.5 & 16.5 & 16.5 & 16.6 & 17.9 & 17.8 \\
\bottomrule
\end{tabular}}
\end{table*}

\section{Covariate routing and native multivariate forecasting}
\label{app:covmv}
These two ablations support \S5.4: the corrector's gain is decoupled from how a
covariate-capable or multivariate-capable backbone uses its richer input path.

\paragraph{Covariates in the backbone vs.\ the corrector.} Table~\ref{tab:covroute} feeds known
covariates to the frozen backbone (\textsc{Raw}$+$cov) instead of routing them to the corrector. Toto
ingests them harmlessly ($\Delta\le0$ on every dataset) while Moirai-2.0 misuses them, degrading
\textsc{Raw} by up to $45.8\%$ (\textsc{rossmann}); neither backbone turns covariate access into a real
gain. Routing the same covariates to the corrector instead yields large gains, and \method{} reaches
comparable accuracy whether or not the backbone was given the covariates---recovering even where the
cov-fed backbone degraded its own \textsc{Raw} (Moirai-2.0 \textsc{rossmann}, $31.1{\to}13.8$
vs.\ $11.8$ on the univariate backbone). The corrector's gain is therefore largely independent of how,
or whether, the backbone ingests covariates; the lone volatile \textsc{epf} series is the one
exception. Timer and Chronos are univariate and cannot ingest covariates.

\begin{table*}[t]
\centering \footnotesize \setlength{\tabcolsep}{8pt}
\caption{\textbf{Covariates belong in the corrector, not the backbone} (DeepSeek-3.4-pro backend, test
\wmape{}$\times100$, three-split mean, seed 42). \textsc{Raw} is the univariate backbone;
\textsc{Raw}$+$cov feeds covariates to the frozen backbone, with $\Delta$ the resulting change in
\textsc{Raw} (positive $=$ worse). \method{} (uni) and \method{} (cov) place the corrector on the
univariate and the cov-fed backbone respectively (the two \colorbox{ourshl}{gray} columns; \textbf{bold}
$=$ beats both \textsc{Raw} and \textsc{Raw}$+$cov in that row, the Table~\ref{tab:master} rule). Feeding
covariates to the backbone is at best
neutral (Toto, $\Delta\le0$ throughout) and often harmful (Moirai-2.0), whereas routing them to the
corrector yields large gains; the two \method{} columns match across the retail panels, so the
corrector's gain is decoupled from how the backbone routes covariates (\textsc{epf} excepted). Timer
and Chronos are univariate and cannot ingest covariates; Moirai covariate injection was run
single-split only; \textsc{bizitobs} has no covariates. \textsc{m5} is evaluated on this grid's own
series basis, so its \textsc{Raw} differs slightly from Table~\ref{tab:master}; compare within the
table.}
\label{tab:covroute}
\begin{tabular}{llccc>{\columncolor{ourshl}}c>{\columncolor{ourshl}}c}
\toprule
BB & Dataset & \textsc{Raw} & \textsc{Raw}$+$cov & $\Delta$ & \method{} (uni) & \method{} (cov) \\
\midrule
\multirow{5}{*}{Toto} & epf      & 55.8 & 52.6 & $-5.7\%$  & \textbf{39.3} & \textbf{50.9} \\
                      & rossmann & 22.2 & 22.1 & $-0.7\%$  & \textbf{13.5} & \textbf{13.6} \\
                      & rohlik   & 28.2 & 27.9 & $-0.9\%$  & \textbf{25.0} & \textbf{24.8} \\
                      & favorita & 17.4 & 17.4 & $-0.1\%$  & 17.6 & 17.6 \\
                      & m5       & 50.6 & 49.3 & $-2.5\%$  & 50.6 & 49.4 \\
\midrule
\multirow{5}{*}{Moirai-2} & epf      & 55.0 & 54.5 & $-0.9\%$  & \textbf{46.5} & \textbf{42.9} \\
                          & rossmann & 21.4 & 31.1 & $+45.8\%$ & \textbf{11.8} & \textbf{13.8} \\
                          & rohlik   & 34.9 & 35.3 & $+1.0\%$  & \textbf{32.3} & \textbf{31.9} \\
                          & favorita & 16.5 & 18.9 & $+14.5\%$ &  \textbf{16.5} & 18.2 \\
                          & m5       & 49.9 & 49.8 & $-0.2\%$  & 50.0 & 50.2 \\
\bottomrule
\end{tabular}
\end{table*}

\paragraph{Native multivariate forecasting.} \textsc{bizitobs} is the one panel with several jointly
forecastable channels. Table~\ref{tab:mv} compares per-channel (uni) with joint (mv) backbone
forecasting on the two natively multivariate backbones. Joint forecasting improves Toto's any-variate
\textsc{Raw} and badly degrades the small Moirai-2.0, but the corrector behaves identically on either
forecast (the same small lift in both the uni and mv columns), so the uni/mv gap is set by the
backbone, not the wrapper. Timer and Chronos are univariate; Moirai is multivariate-capable but
not run here. Importantly, native multivariate forecasting changes the backbone, not the wrapper.
On \textsc{bizitobs}, the only panel where several channels can be forecast jointly, native
multivariate forecasting is strongly backbone-dependent: it improves Toto's raw forecast but degrades
the smaller Moirai-2.0 model (App.~Table~\ref{tab:mv}). 
The correction layer behaves similarly on
either raw forecast. Thus the univariate--multivariate gap is mainly a property of the backbone's
input path, whereas \method{} remains a black-box residual correction layer that applies after either
choice.

\begin{table}[h]
\centering \footnotesize \setlength{\tabcolsep}{6pt}
\caption{\textbf{Native multivariate forecasting on \textsc{bizitobs}} (DeepSeek-3.4-pro backend, test
\wmape{}$\times100$, three-split mean, seed 42). uni $=$ per-channel, mv $=$ joint forecast. \method{}
shows a similar lift over \textsc{Raw} in both columns; the uni/mv difference comes from the
backbone's joint-forecast quality.}
\label{tab:mv}
\begin{tabular}{lcccc}
\toprule
& \multicolumn{2}{c}{\textsc{Raw}} & \multicolumn{2}{c}{\method} \\
\cmidrule(lr){2-3}\cmidrule(lr){4-5}
BB & uni & mv & uni & mv \\
\midrule
Toto     & 39.0 & 38.3 & 41.5 & 39.6 \\
Moirai-2 & 27.1 & 39.2 & 26.2 & 38.8 \\
\bottomrule
\end{tabular}
\end{table}

\section{LLM Roles and Prompts}
\label{app:prompts}
\method{} calls a single LLM provider in four roles: a fast tier runs the \textsc{Planner}, a
reasoning tier runs the \textsc{Specifier} and \textsc{Reflection}, and a \textsc{Base Proposer}
invents new base columns. The legacy \textsc{Reranker} is disabled---the merged candidate pool is
gated deterministically by the source-blind rule of \S3.3, not by an LLM. Across a run
(three rounds) the LLM is queried about a dozen times (${\sim}4$ per round): each exploration round
runs a \textsc{Planner} that proposes the round's directions, a \textsc{Specifier} per direction, a
\textsc{Base Proposer}, and a \textsc{Reflection} call only when fewer than three candidates pass the
gate. Every system prompt is parameterized by a per-dataset domain preamble (the dataset name, a
one-line domain hint, and a role-specific note). We show each role's system and user templates below;
run-time inputs appear as \texttt{\{placeholders\}} and the model must reply in strict JSON.

\paragraph{Planner (fast tier).} \emph{Inputs:} round index, validation-\wmape{} trajectory,
per-family coverage, under-served families, chronic failure subgroups, and rejection fingerprints by
family. \emph{Output:} exactly $K$ exploration directions.
\begin{quote}\small
\emph{System.} ``You are a feature-exploration planner for \texttt{\{domain\}} time-series
forecasting. Each round you see current feature coverage, chronic failure subgroups, recent rejection
fingerprints grouped by family, and the \wmape{} trend over recent rounds. Output $K$ structured
exploration \emph{directions}; each tells the downstream Specifier which family/subgroup to design
for, what type of signal (\texttt{combo}, \texttt{flag}, or both), and which base feats to use. Return
strict JSON \texttt{\{directions:[\{family, target\_subgroup, base\_feat\_pool, spec\_type\_hint,
mechanism\_hint, rationale\}]\}}. Constraints: return exactly $K$ directions; \texttt{family} from the
allowed names; \texttt{base\_feat\_pool} a subset of the supplied feats; do not repeat a
(family,~subgroup) pair; prioritize under-served families and chronic subgroups.''\\[2pt]
\emph{User.} ``Round \texttt{\{r\}}, $K{=}\texttt{\{k\}}$. \wmape{} trajectory (last 5):
\texttt{\{...\}}. Family names, coverage, under-served families, chronic subgroups (top 8), and
rejections by family over the last 3 rounds: \texttt{\{...\}}. Existing dynamic features (top 30, with
$\rho$): \texttt{\{...\}}. Available base feats: \texttt{\{...\}}. Output exactly $K$ directions as
JSON.''
\end{quote}

\paragraph{Specifier (reasoning tier).} \emph{Inputs:} one Planner direction (family, optional
subgroup, base-feat pool, signal-type hint, mechanism hint), up to eight chronic failure cases, and
in-scope rejections to avoid. \emph{Output:} 5--8 named specs (\texttt{combo}/\texttt{flag}/\texttt{code}).
\begin{quote}\small
\emph{System.} ``You are a feature engineer for \texttt{\{domain\}} time-series forecasting. You
receive one exploration direction (family + optional target subgroup + allowed base-feat pool +
mechanism hint). Strictly within that direction, produce 5--8 structured feature specs
(\texttt{combo}, \texttt{flag}, or \texttt{code}). Hard constraints: \texttt{combo}/\texttt{flag} may
use only feats in \texttt{base\_feat\_pool}; \texttt{code} specs may read the target history and the
backbone forecast but may import only \texttt{numpy}, \texttt{math}, and
\texttt{scipy.\{signal,stats,fft,special\}}, with no I/O, \texttt{exec}, \texttt{eval}, \texttt{open},
loops, or \texttt{try}; do not resubmit a spec identical to a rejected one on (op, feat set,
transforms); every spec must pass \texttt{validate\_spec}. Output JSON under \texttt{missing\_features},
each entry a \texttt{combo} (\texttt{feature\_name, feat\_a, feat\_b, op, transform\_a, transform\_b,
post\_transform}), a \texttt{flag} (\texttt{conditions, combine, then, else}), or a \texttt{code} spec
(\texttt{compute\_func\_code}). Guidance: the atomic search already covers plain \texttt{combo}/unary
features, so prioritize \texttt{flag} (multi-condition thresholds the search cannot express) and
\texttt{code} (FFT, autocorrelation, Hurst, windowed statistics); of the 5--8 specs keep
\texttt{combo}~$\le 2$ and \texttt{flag}+\texttt{code}~$\ge 60\%$.''\\[2pt]
\emph{User.} ``family, target\_subgroup, base\_feat\_pool, spec\_type\_hint, mechanism\_hint:
\texttt{\{...\}}. Chronic failure cases (for inspiration): \texttt{\{...\}}. Rejected fingerprints in
this direction (avoid the same mechanism): \texttt{\{...\}}. Produce 5--8 specs as strict JSON under
\texttt{missing\_features}.''
\end{quote}

\paragraph{Reflection (reasoning tier; fires only when ${<}3$ candidates pass the gate in a round).}
\emph{Inputs:} one rejected candidate spec, its rejection reason (\texttt{pair-rho-high} or
\texttt{marginal-rho-low}), and the direction's mechanism hint. \emph{Output:} exactly one rewritten
spec with a different mechanism.
\begin{quote}\small
\emph{System.} ``You are a feature-improvement expert for \texttt{\{domain\}} time-series forecasting.
You receive a candidate rejected by the novelty gate, its rejection reason, and the direction's
mechanism hint. Rewrite one new spec with a \emph{different} mechanism---you must change at least one
of \texttt{transform}, \texttt{feat\_b}, or \texttt{op}; renaming alone does not count. Output JSON
under \texttt{missing\_features} containing exactly one spec.''\\[2pt]
\emph{User.} ``Rejected candidate: \texttt{\{spec\}}. reject\_reason: \texttt{\{...\}}.
mechanism\_hint: \texttt{\{...\}}. Rewrite one spec with a different mechanism.''
\end{quote}

\paragraph{Base Proposer (invents new base columns that join the channel pool and become searchable
next round).} \emph{Inputs:} round index, existing base-feature names, recent residual statistics, and
chronic worst subgroups. \emph{Output:} new univariate base-feature functions.
\begin{quote}\small
\emph{System.} ``You design new base-level features for \texttt{\{domain\}} time-series forecasting
that feed the residual-correction GBDT. Each feature is a Python function
\texttt{compute\_NAME(history, chronos\_pred)} returning one scalar, where \texttt{chronos\_pred} is
the backbone forecast. Allowed: \texttt{numpy} (as \texttt{np}), arithmetic, and simple control flow.
Forbidden: any \texttt{import}, I/O, \texttt{exec}/\texttt{eval}/\texttt{open}, and referencing both
\texttt{history} and \texttt{chronos\_pred} in one function (each base feat must summarize a single
input; cross-source interactions are left to the atomic search). Output strict JSON under
\texttt{proposals}, each \texttt{\{feature\_name, family, rationale, code\}} with \texttt{family} from
the allowed set. Target patterns the existing base set under-represents and keep proposals mutually
orthogonal.''\\[2pt]
\emph{User.} ``Round \texttt{\{r\}}. Generate \texttt{\{n\}} new base features. Existing base names
(avoid duplicates): \texttt{\{...\}}. Recent residual stats: \texttt{\{...\}}. Chronic worst subgroups:
\texttt{\{...\}}. Output strict JSON with a \texttt{proposals} list; each function takes
(\texttt{history}, \texttt{chronos\_pred}) and returns a float.''
\end{quote}

\section{Framework Comparison: With vs.\ Without the LLM}
\label{app:llm-vs-nollm}
Table~\ref{tab:llm-vs-nollm} contrasts the \method{} pipeline with and without the LLM agent at the
component level. The two columns are the \method-noLLM and \method{} columns of Table~\ref{tab:master};
the counts are means per cell over the GPT-5.2 grid (selected $=$ admitted to the corrector). Adding the
compositional search on top of the covariate-only corrector moves accuracy by ${\approx}0$
(better-or-equal on $10/30$ cells, mean $+0.006$ \wmape{}); adding the LLM on top of the search is the
decisive step (mean $-0.037$ \wmape{}, better on $28/30$ cells), and the gain concentrates on the
weak-backbone cells where a semantic signal exists. At the level of accepted features, the LLM also dominates the final corrector: about
three-quarters of admitted features are LLM-proposed, with LLM-combo features the most efficient and
LLM-code features almost always pruned (per-source counts in Table~\ref{tab:source},
broken down by dataset and backbone in \S\ref{app:source}).

\begin{table}[htbp]
\centering
\footnotesize
\caption{Component-level comparison of \method{} with vs.\ without the LLM agent. Counts are per-cell
means over the GPT-5.2 grid.}
\label{tab:llm-vs-nollm}
\setlength{\tabcolsep}{4pt}
\begin{tabular}{@{}>{\raggedright\arraybackslash}p{0.30\columnwidth}
                   >{\raggedright\arraybackslash}p{0.28\columnwidth}
                   >{\raggedright\arraybackslash}p{0.34\columnwidth}@{}}
\toprule
Aspect & no-LLM (search only) & \method{} (full) \\
\midrule
Feature source        & composed channels & $+$ LLM-flag/combo/code \\
Semantic naming       & no                & yes \\
Search guidance       & syntactic UCB     & LLM directions $+$ UCB \\
LLM calls / round     & $0$               & ${\sim}4$ \\
Interpretable names   & composed exprs    & named domain features \\
Selected features     & ${\sim}3.5$ (all atomic) & ${\sim}12.4$ (${\sim}74\%$ LLM) \\
Gain vs.\ previous step & ${\approx}0$ ($10/30$) & $-0.037$ \wmape{} ($28/30$) \\
Where it helps        & composed structure & weak backbone, semantic signal \\
\bottomrule
\end{tabular}
\end{table}

\section{Feature-source anatomy by dataset and backbone}
\label{app:source}
Table~\ref{tab:source} aggregates the per-source \emph{proposed} and \emph{selected} feature counts and
survival rates over the grid (the source for Fig.~3a); Tables~\ref{tab:source-dataset}
and~\ref{tab:source-backbone} break them down by dataset and by backbone (GPT-5.2 fine-tuning grid, pooled
over the two \method{} variants and three splits). Three patterns hold. The selected total rises with
residual headroom: the weak-backbone, signal-rich \textsc{favorita} cells
admit the most features, while the small or saturated \textsc{epf} and \textsc{bizitobs} cells sit at the
floor. LLM-flag features are admitted mainly where the data has explicit
calendar structure (\textsc{favorita}); on \textsc{epf} and \textsc{bizitobs} most
are trimmed. LLM-code features are pruned almost everywhere. The mix is insensitive to the
backbone (Moirai ${\approx}$ Moirai-2.0).

\begin{table}[htbp]
\centering
\footnotesize
\setlength{\tabcolsep}{4pt}
\caption{\textbf{Feature-source anatomy (aggregate).} Per-cell mean \emph{proposed} (P) and
\emph{selected} (S, admitted to the corrector) feature counts by source, with survival rate $S/\text{P}$,
over the fine-tuning grid (both Moirai backbones, three splits). LLM features
supply about three-quarters of the selected set; LLM-code is almost always pruned.}
\label{tab:source}
\begin{tabular}{lccccccc}
\toprule
& \multicolumn{3}{c}{\method{}} & \multicolumn{3}{c}{\method$^{\dagger}$} \\
\cmidrule(lr){2-4}\cmidrule(lr){5-7}
Source & P & S & surv & P & S & surv \\
\midrule
atomic           & $5.7$ & $3.5$ & $61\%$ & $5.7$ & $3.3$ & $58\%$ \\
LLM-flag         & $11.7$ & $4.8$ & $41\%$ & $10.8$ & $4.7$ & $44\%$ \\
LLM-combo        & $5.7$ & $3.9$ & $68\%$ & $6.6$ & $4.3$ & $65\%$ \\
LLM-code         & $2.3$ & $0.2$ & $9\%$ & $2.4$ & $0.2$ & $7\%$ \\
\midrule
Total            & $25.4$ & $12.4$ & $49\%$ & $25.5$ & $12.5$ & $49\%$ \\
LLM share of S   &  & $72\%$ &  &  & $74\%$ &  \\
\bottomrule
\end{tabular}
\end{table}

\begin{table}[htbp]
\centering
\footnotesize
\setlength{\tabcolsep}{3.5pt}
\caption{Feature-source anatomy by dataset (per-cell mean, proposed/selected). $P$, $S$ are the totals;
LLM\%(S) is the LLM share of selected features.}
\label{tab:source-dataset}
\resizebox{\columnwidth}{!}{%
\begin{tabular}{lccccccc}
\toprule
Dataset & atomic & flag & combo & code & $P$ & $S$ & LLM\%(S) \\
\midrule
\textsc{epf}      & $5.5/3.1$ & $9.2/2.8$  & $6.8/4.1$ & $2.6/0.0$ & $24.1$ & $10.0$ & $69\%$ \\
\textsc{rossmann} & $5.7/3.3$ & $11.2/4.2$ & $6.6/3.8$ & $1.6/0.1$ & $25.1$ & $11.5$ & $71\%$ \\
\textsc{bizitobs} & $5.3/2.5$ & $9.2/2.0$  & $7.8/4.9$ & $2.3/0.6$ & $24.6$ & $10.0$ & $75\%$ \\
\textsc{favorita} & $6.0/4.0$ & $12.1/7.2$ & $6.4/4.8$ & $2.9/0.0$ & $27.4$ & $16.0$ & $75\%$ \\
\bottomrule
\end{tabular}}
\end{table}

\begin{table}[htbp]
\centering
\footnotesize
\setlength{\tabcolsep}{3.5pt}
\caption{Feature-source anatomy by backbone (per-cell mean, proposed/selected).}
\label{tab:source-backbone}
\resizebox{\columnwidth}{!}{%
\begin{tabular}{lccccccc}
\toprule
Backbone & atomic & flag & combo & code & $P$ & $S$ & LLM\%(S) \\
\midrule
Moirai & $5.8/3.4$ & $11.3/4.4$ & $6.2/4.7$ & $2.3/0.2$ & $25.6$ & $12.6$ & $73\%$ \\
Moirai-2.0 & $5.6/3.4$ & $11.2/5.1$ & $6.1/3.5$ & $2.4/0.2$ & $25.3$ & $12.2$ & $72\%$ \\
\bottomrule
\end{tabular}}
\end{table}

\section{Representative Accepted Features}
\label{app:examples}
The examples below illustrate what each source contributes. They are drawn from the per-feature audit
logs of the DeepSeek \method$^{\dagger}$ runs (rolling split $8$, seed $42$), which record every
candidate's acceptance correlation $\rho$ and its split gain in the final corrector. ``Share'' is the
feature's fraction of its cell's total corrector gain; gains are target-scale dependent, so shares
compare only within a cell.

\paragraph{Atomic.}
Compositions the search assembled from raw channels; the recorded name encodes the expression, \eg{}
\texttt{binary\_safe\_div\_log-cov\_Promo\_softplus-dow}. That feature,
$\mathrm{safe\_div}(\log(\texttt{Promo}),\,\mathrm{softplus}(\texttt{dow}))$ (\textsc{rossmann},
Moirai-2; $\rho{=}0.61$, share $50\%$), is a promotion-by-weekday interaction, and the identical spec
is also accepted in the decoupled \method{} run of the same cell.
$\min(\mathrm{softplus}(\texttt{hour}),\,\texttt{is\_weekend})$ (\textsc{bizitobs}, Timer;
$\rho{=}{-}0.78$, share $48\%$) gates the hourly IT-traffic residual by weekend and time of day.
$\min(\log(\texttt{holiday}),\,\tanh(\texttt{is\_weekend}))$ (\textsc{rohlik}) is selected under four
of the five backbones; for a tree corrector it reduces to a re-derived holiday indicator, typical of
how the search re-expresses information the schema already carries rather than adding semantics.

\paragraph{LLM-combo.}
Named combinations with a stated mechanism. \texttt{renewable\_penetration}, the renewable-generation
forecast over the load forecast (\textsc{epf}, Moirai; $\rho{=}{-}0.81$, share $81\%$), was proposed
because ``values above 1.0 directly signal surplus conditions \dots which modulates price
volatility''; a renewables-over-load variant is selected on all five \textsc{epf} backbones.
\texttt{baseline\_mean\_x\_open}, the context mean $\times$ \texttt{Open} (\textsc{rossmann},
Moirai; share $88\%$), is the expected sales level when the store is open. \texttt{event\_ratio},
the last value over the context mean (\textsc{favorita}, Chronos; share $37\%$), was proposed to flag
``sudden drops typical of earthquake shocks''.

\paragraph{LLM-flag.}
The LLM proposes the regime and its channels; the numeric thresholds are pool quantiles filled in by
the enumerator. On \textsc{rossmann} (Moirai-2; share $14\%$) an anomaly direction targeting abnormal
promotion days yields
$\mathbf{1}[\texttt{ctx\_std}{>}q_{75} \wedge \texttt{Promo}{>}0 \wedge \texttt{Open}{\geq}1]$, an
open, high-volatility store running a promotion. On \textsc{bizitobs} (Toto; $\rho{=}{-}0.61$, share
$18\%$), $\mathbf{1}[\texttt{is\_weekend} \vee h{<}6]$ marks the low-load regime of the API-traffic
series. On \textsc{favorita} (Moirai),
$\mathbf{1}[\texttt{onpromotion}{\leq}0 \wedge \texttt{holiday}{\leq}0]$ separates quiet baseline days
from event days; holiday-conditioned flags are selected on four of the five \textsc{favorita}
backbones.

\paragraph{LLM-code.}
Selected code features are rare (Table~\ref{tab:source}), and the survivors compute statistics the
grammar cannot express. \texttt{trend\_r\_squared} (\textsc{favorita}, Toto; share $55\%$) fits a line
to the last $28$ days and returns its $R^2$, proposed as ``how strongly a linear model explains recent
sales, complementary to slope magnitude''. \texttt{mk\_trend\_tau} (\textsc{m5}, Timer; share $28\%$)
is Kendall's $\tau$ of demand against time, a monotone-trend strength robust to outliers; its accepted
source is
{\small
\begin{verbatim}
def compute_mk_trend_tau(history, chronos_pred):
    import numpy as np
    from scipy import stats
    if len(history) < 5:
        return 0.0
    tau, p = stats.kendalltau(
        np.arange(len(history)), history)
    return float(tau) if not np.isnan(tau) else 0.0
\end{verbatim}}

\section{Feature transferability (faithful replay)}
\label{app:transfer}
We test whether features learned on one backtest window stay useful on a later one. For an ordered split
pair $S\!\to\!T$ we load the accepted feature specifications (combinations and synthesized code) from the
source cell, force-carry them into the target split's accepted set, re-materialize the code features on
the target data, and refit and evaluate the corrector with no new proposals (LLM and search off,
validation gating on). Two references on the same target bound the effect: \emph{self}, the target's own
accepted features replayed back (a faithfulness upper bound that should reproduce native \method{}), and
\emph{cov}, a covariate-only corrector (the no-transfer lower bound). The grid is two datasets
(\textsc{epf}, \textsc{bizitobs}) $\times$ five backbones $\times$ three split
pairs ($6\!\to\!7$, $7\!\to\!8$, $6\!\to\!8$) $\times$ \{cov, self, xfer\} $=90$ cells, all completing.
The self column at split $S$ reproduces native \method{} closely, so the replay is faithful.
Table~\ref{tab:transfer-summary} summarizes the $30$ source-cell triples and
Table~\ref{tab:transfer} lists every cell.

Three findings hold. Transferred features match the target's own at the median (rel.\ $+0.2\%$, $17/30$
within $\pm5\%$); the mean is pulled up only by a heavy tail. Distance decay is mild: the median for
the two-split jump ($6\!\to\!8$) is only a couple of percent above the adjacent-split median. The $7/30$
catastrophic transfers (worse than self by $>15\%$) concentrate on the single-series \textsc{epf} and on
Moirai backbones---the shift-sensitive corner of \S5.2---while the multivariate
\textsc{bizitobs} panel transfers with median $+0.0\%$. Low feature-name overlap across splits therefore
reflects equivalent-feature redundancy, not non-transferability.

\begin{table}[h]
\centering
\footnotesize
\setlength{\tabcolsep}{4pt}
\caption{Transferability summary over the $30$ source-cell triples. ``rel.'' is the relative test
\wmape{} gap to the reference (negative favors transfer).}
\label{tab:transfer-summary}
\resizebox{\columnwidth}{!}{%
\begin{tabular}{lll}
\toprule
Question & Metric & Result \\
\midrule
xfer vs.\ self (upper bound) & median rel. & $+0.2\%$; $17/30$ within $\pm5\%$ \\
xfer vs.\ cov (lower bound)  & median rel. & $+0.0\%$; $12/30$ beat cov \\
distance decay               & adj.\ vs.\ far & $+0.0\%$ vs.\ $+1.7\%$ (mild) \\
catastrophic ($>15\%$)       & count & $7/30$, \textsc{epf}/Moirai-dominated \\
by dataset                   & median rel.\ to self & \textsc{epf} $+5.6\%$, \textsc{bizitobs} $+0.0\%$ \\
\bottomrule
\end{tabular}}
\end{table}

\begin{table}[h]
\centering
\footnotesize
\setlength{\tabcolsep}{3.5pt}
\caption{Per-cell faithful replay (test \wmape{}$\times100$, lower is better). \emph{cov}: covariate-only
target corrector (no-transfer lower bound); \emph{self}: target's own accepted features replayed
(faithfulness upper bound); \emph{xfer}: source-split features replayed on the target. $\Delta_{\text{s}}$,
$\Delta_{\text{c}}$ are the relative gaps of xfer to self and to cov in \%. Pairs $7\!\to\!8$ and
$6\!\to\!8$ share the same target (split~$8$), so their cov/self match.}
\label{tab:transfer}

\begin{tabular}{llrrrrr}
\toprule
BB & pair & cov & self & xfer & $\Delta_{\text{s}}$ & $\Delta_{\text{c}}$ \\
\midrule
\multicolumn{7}{l}{\textit{\textsc{epf}}}\\
tim & $6{\to}7$ & $62.8$ & $63.9$ & $62.8$ & $-1.7$ & $+0.0$ \\
tim & $7{\to}8$ & $23.9$ & $23.2$ & $23.4$ & $+0.9$ & $-2.4$ \\
tim & $6{\to}8$ & $23.9$ & $23.2$ & $23.8$ & $+2.9$ & $-0.5$ \\
chr & $6{\to}7$ & $62.3$ & $62.6$ & $66.1$ & $+5.6$ & $+6.0$ \\
chr & $7{\to}8$ & $26.3$ & $26.3$ & $25.9$ & $-1.4$ & $-1.6$ \\
chr & $6{\to}8$ & $26.3$ & $26.3$ & $26.4$ & $+0.5$ & $+0.3$ \\
moi & $6{\to}7$ & $59.0$ & $65.9$ & $115.3$ & $+75.0$ & $+95.5$ \\
moi & $7{\to}8$ & $25.0$ & $19.8$ & $22.2$ & $+11.8$ & $-11.5$ \\
moi & $6{\to}8$ & $25.0$ & $19.8$ & $47.5$ & $+139.5$ & $+89.6$ \\
toto & $6{\to}7$ & $61.5$ & $61.2$ & $63.1$ & $+3.0$ & $+2.5$ \\
toto & $7{\to}8$ & $23.7$ & $19.7$ & $21.9$ & $+11.1$ & $-7.4$ \\
toto & $6{\to}8$ & $23.7$ & $19.7$ & $22.8$ & $+15.7$ & $-3.6$ \\
moi$_2$ & $6{\to}7$ & $62.0$ & $90.2$ & $61.1$ & $-32.3$ & $-1.6$ \\
moi$_2$ & $7{\to}8$ & $24.3$ & $19.5$ & $55.1$ & $+182.0$ & $+126.8$ \\
moi$_2$ & $6{\to}8$ & $24.3$ & $19.5$ & $23.3$ & $+19.2$ & $-4.1$ \\
\midrule
\multicolumn{7}{l}{\textit{\textsc{bizitobs}}}\\
tim & $6{\to}7$ & $31.8$ & $33.3$ & $31.1$ & $-6.7$ & $-2.3$ \\
tim & $7{\to}8$ & $30.5$ & $31.7$ & $30.8$ & $-3.0$ & $+0.9$ \\
tim & $6{\to}8$ & $30.5$ & $31.7$ & $30.2$ & $-4.8$ & $-0.9$ \\
chr & $6{\to}7$ & $22.6$ & $22.6$ & $22.6$ & $+0.0$ & $+0.0$ \\
chr & $7{\to}8$ & $21.7$ & $21.7$ & $21.7$ & $+0.0$ & $+0.0$ \\
chr & $6{\to}8$ & $21.7$ & $21.7$ & $21.7$ & $+0.0$ & $+0.0$ \\
moi & $6{\to}7$ & $36.2$ & $38.0$ & $36.5$ & $-4.0$ & $+0.8$ \\
moi & $7{\to}8$ & $32.1$ & $32.2$ & $40.4$ & $+25.2$ & $+25.8$ \\
moi & $6{\to}8$ & $32.1$ & $32.2$ & $35.7$ & $+10.9$ & $+11.4$ \\
toto & $6{\to}7$ & $33.8$ & $31.7$ & $31.6$ & $-0.2$ & $-6.4$ \\
toto & $7{\to}8$ & $30.3$ & $31.3$ & $30.1$ & $-3.9$ & $-0.5$ \\
toto & $6{\to}8$ & $30.3$ & $31.3$ & $30.6$ & $-2.3$ & $+1.2$ \\
moi$_2$ & $6{\to}7$ & $24.8$ & $24.8$ & $29.0$ & $+16.6$ & $+16.6$ \\
moi$_2$ & $7{\to}8$ & $23.9$ & $23.9$ & $23.9$ & $+0.0$ & $+0.0$ \\
moi$_2$ & $6{\to}8$ & $23.9$ & $23.9$ & $23.9$ & $+0.0$ & $+0.0$ \\
\bottomrule
\end{tabular}
\end{table}

\section{Feature-budget sweep}
\label{app:budget}
Table~\ref{tab:budget} sweeps the feature cap $K$ over the $30$ GPT-5.2 cells (each a
three-split mean), backing Fig.~2b and the budget claim of \S5.1. The
externals here are the honest budget-$50$ sweep (\texttt{grid\_allbb\_uncap}), so their $K{\geq}10$
columns are genuinely re-generated rather than repeats of the small cap. The
win rate over \textsc{Raw} is constant at $80\%$ across every $K$: the budget never changes which cells
\method{} wins, only by how much. The margin over the best external system instead widens
with $K$ and then saturates at $K{=}20$ ($K{=}20/50/$all are nearly identical), so $K{=}20$ is the
practical operating point. Split by residual headroom, the budget amplifies an existing gap rather than
creating one: on raw-weak cells (external moves \textsc{Raw} by ${\geq}2\%$) the mean lift over
\textsc{Raw} reaches $-0.131$ at $K{=}20$ ($20/20$ cells), whereas on raw-strong cells it sits near
$+0.008$ and a larger cap, if anything, fits more noise.

\begin{table}[h]
\centering
\footnotesize
\setlength{\tabcolsep}{5pt}
\caption{Feature-budget sweep (GPT-5.2, $30$ cells, three-split mean). Win rates and mean differences in
test \wmape{} for \method{} (decoupled) vs.\ \textsc{Raw} and vs.\ the best external system (negative
mean favors \method{}).}
\label{tab:budget}
\resizebox{\columnwidth}{!}{%
\begin{tabular}{ccccc}
\toprule
$K$ & \method$<$\textsc{Raw} & mean($-$\textsc{Raw}) & \method$<$best-ext & mean($-$ext) \\
\midrule
$5$   & $80\%$ & $-0.067$ & $60\%$ & $-0.020$ \\
$10$  & $80\%$ & $-0.073$ & $73\%$ & $-0.025$ \\
$20$  & $80\%$ & $-0.085$ & $73\%$ & $-0.038$ \\
$50$  & $80\%$ & $-0.085$ & $73\%$ & $-0.039$ \\
all   & $80\%$ & $-0.085$ & $73\%$ & $-0.036$ \\
\bottomrule
\end{tabular}}
\end{table}

\section{Full per-cell results}
\label{app:fulltables}
\begin{figure}[t]
\centering
\includegraphics[width=0.86\columnwidth]{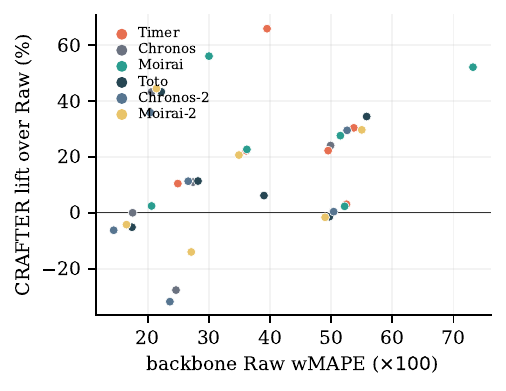}
\caption{\textbf{Gain vs.\ backbone competence (per cell, GPT-5.2).} Each point is one
(dataset, backbone) cell: $x$ is the frozen backbone's Raw \wmape{}$\times100$, $y$ is \method's lift
over Raw. The gain trends positive with backbone error (more exploitable residual) but is noisy across
datasets; the per-backbone means (Fig.~2a) summarize the trend. Residual headroom, not
dataset or backbone identity, orders the gain: points on the high-error right are consistently corrected,
while saturated low-error points on the left cluster near zero or below, the per-cell form of the
headroom condition (\S5.2).}
\label{fig:headroom}
\end{figure}

Table~\ref{tab:master} gives every cell of the GPT-5.2 main grid (the source for Fig.~2a);
Table~\ref{tab:deepseek} is the DeepSeek-3.4-pro reproduction. Figure~\ref{fig:headroom} plots the
per-cell gain against backbone competence: the lift over \textsc{Raw} trends positive with the
backbone's own error but is noisy, which is why the body reports the cleaner per-backbone aggregate
(Fig.~2a) rather than the raw scatter. A single quantity---how much
exploitable residual the frozen forecast leaves---orders the gain: the wrapper is net-positive on the
weak, high-error cells to the right and neutral-to-negative on the saturated, low-error cells to the
left, with no separate dependence on dataset or on which foundation model produced the point. The
per-cell scatter is the headroom condition of \S5.2 before averaging.

\begin{table*}[t]
\centering \footnotesize \setlength{\tabcolsep}{4pt}
\caption{\textbf{Main results.} \method{} (two rightmost, \colorbox{ourshl}{gray} columns) vs.\ the frozen
backbone (\textsc{Raw}), a head-only backbone fine-tune (\textsc{FT}), a covariate-only corrector
(\textsc{Cov}), the no-LLM corrector (noLLM), and three external feature-engineering systems, on the
$6{\times}6{=}36$-cell grid (test \wmape{}$\times100$, lower is better). Every feature-engineering method
uses the \textbf{same GPT-5.2 backend} and harness, so only the feature \emph{generator} differs. The
noLLM, external, and \method{} columns average three rolling splits $\times$ three seeds; \textsc{Raw},
\textsc{Cov}, and \textsc{FT} involve no LLM or search randomness and are single-run on the same splits.
Per-cell seed variability is quantified on the DeepSeek companion grid (Table~\ref{tab:deepseek}
caption; App.~\ref{app:protocol}).
\textsc{chr} is Chronos (univariate); \textsc{chr}$_2$ is the covariate-aware Chronos-2, reported
single-seed (42) on the same splits.
\method{} is the full system; \method$^{\dagger}$ adds source coupling (\S5.2 of the main paper).
\textbf{Bold} $=$ a \method{} variant beats \emph{every} non-fine-tuned baseline in that row (both may
bold); if neither does, the row's best method is bold. A star ($^{*}$) marks a bold \method{} variant
that \emph{also} beats \textsc{FT}. \textsc{Raw}, \textsc{Cov} and \textsc{FT} are LLM-independent and
shared with Table~\ref{tab:deepseek}; the stronger full-parameter Moirai fine-tune is in
Table~\ref{tab:ft}.}
\label{tab:master}
\resizebox{\textwidth}{!}{%
\begin{tabular}{llccccccc>{\columncolor{ourshl}}c>{\columncolor{ourshl}}c}
\toprule
Dataset & BB & \textsc{Raw} & \textsc{FT} & \textsc{Cov} & {\scriptsize noLLM} & \textsc{TSFresh} & CAAFE & LLM-FE & \method & \method$^{\dagger}$ \\
\midrule
\multirow{6}{*}{epf} & tim & 53.7 & 56.9 & 46.2 & 48.9 & 44.7 & 49.9 & 50.5 & \textbf{37.4}\textsuperscript{*} & \textbf{37.1}\textsuperscript{*} \\
 & chr & 49.9 & 52.0 & 45.4 & 42.2 & 46.5 & 46.5 & 47.5 & \textbf{37.9}\textsuperscript{*} & \textbf{31.6}\textsuperscript{*} \\
 & moi & 73.2 & 71.2 & 60.7 & 62.7 & 60.3 & 62.0 & 58.7 & \textbf{35.1}\textsuperscript{*} & \textbf{35.3}\textsuperscript{*} \\
 & toto & 55.8 & 55.8 & 49.8 & 49.9 & 49.9 & 45.1 & 46.2 & \textbf{36.6}\textsuperscript{*} & \textbf{32.7}\textsuperscript{*} \\
 & chr$_2$ & 52.6 & 55.1 & 43.7 & 41.3 & 50.6 & 42.4 & 44.0 & \textbf{37.1}\textsuperscript{*} & \textbf{32.2}\textsuperscript{*} \\
 & moi$_2$ & 55.0 & 55.0 & 41.6 & 42.1 & 42.3 & 43.9 & 42.0 & \textbf{38.7}\textsuperscript{*} & \textbf{40.0}\textsuperscript{*} \\
\midrule
\multirow{6}{*}{rossmann} & tim & 39.5 & 39.5 & 16.7 & 16.7 & 13.9 & 17.2 & 17.2 & \textbf{13.5}\textsuperscript{*} & \textbf{13.7}\textsuperscript{*} \\
 & chr & 20.6 & 21.3 & 11.9 & 11.9 & 12.2 & 12.0 & 12.0 & \textbf{11.7}\textsuperscript{*} & \textbf{11.7}\textsuperscript{*} \\
 & moi & 30.0 & 29.8 & 24.7 & 26.4 & 24.8 & 24.8 & 24.8 & \textbf{13.2}\textsuperscript{*} & \textbf{13.1}\textsuperscript{*} \\
 & toto & 22.2 & 22.2 & 14.5 & 14.5 & 13.9 & 14.4 & 14.3 & \textbf{12.6}\textsuperscript{*} & \textbf{12.5}\textsuperscript{*} \\
 & chr$_2$ & 20.4 & 21.1 & 15.9 & 15.8 & 15.5 & 15.9 & 15.9 & \textbf{13.1}\textsuperscript{*} & \textbf{12.8}\textsuperscript{*} \\
 & moi$_2$ & 21.4 & 21.4 & 12.6 & 12.7 & 12.8 & 12.6 & 12.6 & \textbf{11.9}\textsuperscript{*} & \textbf{12.0}\textsuperscript{*} \\
\midrule
\multirow{6}{*}{rohlik} & tim & 36.1 & 30.3 & 33.4 & 33.4 & 33.3 & 32.9 & 32.8 & \textbf{28.1}\textsuperscript{*} & \textbf{28.1}\textsuperscript{*} \\
 & chr & 27.4 & 27.2 & 24.7 & 24.6 & 24.6 & 24.7 & 24.6 & \textbf{24.4}\textsuperscript{*} & \textbf{24.4}\textsuperscript{*} \\
 & moi & 36.2 & 31.2 & 33.5 & 33.5 & 33.4 & 32.8 & 32.7 & \textbf{28.0}\textsuperscript{*} & \textbf{28.0}\textsuperscript{*} \\
 & toto & 28.2 & 28.2 & 25.1 & 25.1 & 25.1 & 25.2 & 25.2 & \textbf{25.0}\textsuperscript{*} & \textbf{24.9}\textsuperscript{*} \\
 & chr$_2$ & 26.6 & 27.1 & 23.8 & 23.8 & 23.8 & 23.9 & 23.9 & \textbf{23.6}\textsuperscript{*} & \textbf{23.6}\textsuperscript{*} \\
 & moi$_2$ & 34.9 & 29.2 & 32.2 & 32.3 & 32.1 & 31.9 & 31.8 & \textbf{27.7}\textsuperscript{*} & \textbf{27.7}\textsuperscript{*} \\
\midrule
\multirow{6}{*}{bizitobs} & tim & 49.5 & 51.9 & 42.4 & 43.2 & 42.5 & 43.6 & 43.6 & \textbf{38.5}\textsuperscript{*} & \textbf{37.4}\textsuperscript{*} \\
 & chr & \textbf{24.6} & 25.0 & 25.7 & 34.1 & 25.7 & 25.7 & 25.7 & 31.4 & 30.3 \\
 & moi & 51.5 & 50.9 & 43.5 & 43.9 & 43.0 & 45.1 & 44.2 & \textbf{37.3}\textsuperscript{*} & \textbf{37.5}\textsuperscript{*} \\
 & toto & 39.0 & 39.0 & 40.9 & 37.6 & 38.9 & 40.8 & 40.8 & \textbf{36.6}\textsuperscript{*} & \textbf{35.6}\textsuperscript{*} \\
 & chr$_2$ & \textbf{23.6} & 25.3 & 31.0 & 30.9 & 24.4 & 24.4 & 24.4 & 31.1 & 31.4 \\
 & moi$_2$ & \textbf{27.1} & 27.1 & 27.7 & 29.6 & 27.9 & 28.0 & 28.0 & 30.9 & 29.6 \\
\midrule
\multirow{6}{*}{favorita} & tim & 24.9 & 23.2 & 21.8 & 23.1 & \textbf{21.5} & 21.7 & 21.6 & 22.3 & 22.7 \\
 & chr & 17.5 & 17.4 & 17.6 & 18.0 & 17.4 & 17.6 & 17.6 & 17.5 & \textbf{17.4} \\
 & moi & 20.6 & 20.6 & 19.8 & 20.3 & 19.9 & 19.8 & \textbf{19.6} & 20.1 & 20.3 \\
 & toto & \textbf{17.4} & 17.4 & 17.7 & 18.2 & 17.4 & 17.6 & 17.6 & 18.3 & 18.2 \\
 & chr$_2$ & \textbf{14.4} & 14.6 & 15.2 & 15.3 & 14.5 & 14.7 & 14.6 & 15.3 & 15.9 \\
 & moi$_2$ & \textbf{16.5} & 16.5 & 16.6 & 17.4 & 16.5 & 16.6 & 16.6 & 17.2 & 17.2 \\
\midrule
\multirow{6}{*}{m5} & tim & 52.5 & 51.4 & 51.0 & 51.0 & 51.0 & 51.0 & 51.0 & \textbf{50.9}\textsuperscript{*} & \textbf{50.9}\textsuperscript{*} \\
 & chr & 50.3 & 49.9 & 50.3 & 50.7 & 50.3 & 50.3 & 50.3 & \textbf{50.1} & \textbf{50.2} \\
 & moi & 52.2 & 52.2 & 52.1 & 52.3 & 52.1 & 52.1 & 52.1 & \textbf{51.0}\textsuperscript{*} & \textbf{51.6}\textsuperscript{*} \\
 & toto & \textbf{49.7} & 49.7 & 50.1 & 50.5 & 50.1 & 50.1 & 50.0 & 50.4 & 50.4 \\
 & chr$_2$ & 50.4 & 49.5 & 51.2 & 51.2 & 50.6 & 50.6 & 50.5 & \textbf{50.2} & \textbf{50.3} \\
 & moi$_2$ & \textbf{49.0} & 49.0 & 49.3 & 49.9 & 49.3 & 49.3 & 49.3 & 49.8 & 49.8 \\
\bottomrule
\end{tabular}}
\end{table*}

\begin{table*}[t]
\centering \footnotesize \setlength{\tabcolsep}{4pt}
\caption{\textbf{DeepSeek-3.4-pro full grid} (cross-LLM reproduction; test \wmape{}$\times100$, $6{\times}6{=}36$ cells, lower is better). Companion to the GPT-5.2 Table~\ref{tab:master} with the same columns; the
per-backbone summary is in Table~1 of the main paper. \textsc{Raw}, \textsc{Cov} and \textsc{FT} are
LLM-independent and shared with Table~\ref{tab:master}. noLLM, \method{} and the externals are the
DeepSeek runs; \method$^{\dagger}$ is the DeepSeek source-coupling variant (3 splits, single-seed). \textbf{Bold}
and star ($^{*}$) follow Table~\ref{tab:master}. Across the three seeds the median per-cell std of the
split-averaged value is $0.02$ (noLLM), $0.01$ (\textsc{TSFresh}), $0.05$ (CAAFE), $0.08$ (LLM-FE), and
$0.49$ (\method); \textsc{Raw} is seed-deterministic, and the largest \method{} stds all sit on
\textsc{epf} and \textsc{bizitobs}. The \textsc{chr}$_2$ (Chronos-2) rows are single-seed (42) on the
same three splits, matching how Chronos-2 is reported in Table~\ref{tab:master}.}
\label{tab:deepseek}
\resizebox{\textwidth}{!}{%
\begin{tabular}{llccccccc>{\columncolor{ourshl}}c>{\columncolor{ourshl}}c}
\toprule
Dataset & BB & \textsc{Raw} & \textsc{FT} & \textsc{Cov} & {\scriptsize noLLM} & \textsc{TSFresh} & CAAFE & LLM-FE & \method & \method$^{\dagger}$ \\
\midrule
\multirow{6}{*}{epf} & tim & 53.7 & 56.9 & 46.2 & 43.7 & 44.7 & 50.1 & 53.4 & \textbf{43.4}\textsuperscript{*} & \textbf{31.3}\textsuperscript{*} \\
 & chr & 49.9 & 52.0 & 45.4 & \textbf{42.6} & 45.2 & 47.6 & 45.7 & 48.5 & 43.9 \\
 & moi & 73.2 & 71.2 & 60.7 & 57.2 & 56.6 & 60.7 & 57.5 & 59.2 & \textbf{28.6}\textsuperscript{*} \\
 & toto & 55.8 & 55.8 & 49.8 & 50.3 & 49.9 & 45.6 & 45.9 & \textbf{43.9}\textsuperscript{*} & \textbf{40.9}\textsuperscript{*} \\
 & chr$_2$ & 52.6 & 55.1 & 43.7 & 47.0 & 42.9 & 43.6 & 44.3 & 46.8 & \textbf{37.2}\textsuperscript{*} \\
 & moi$_2$ & 55.0 & 55.0 & 41.6 & 42.4 & 42.3 & 45.1 & 42.8 & 44.4 & \textbf{36.0}\textsuperscript{*} \\
\midrule
\multirow{6}{*}{rossmann} & tim & 39.5 & 39.5 & 16.7 & 17.0 & 13.8 & 17.4 & 17.4 & 14.2 & \textbf{13.2}\textsuperscript{*} \\
 & chr & 20.6 & 21.3 & 11.9 & 11.5 & 12.3 & 11.9 & 11.9 & \textbf{11.5}\textsuperscript{*} & \textbf{11.5}\textsuperscript{*} \\
 & moi & 30.0 & 29.8 & 24.7 & 25.0 & 24.8 & 24.8 & 24.8 & \textbf{23.9}\textsuperscript{*} & \textbf{13.2}\textsuperscript{*} \\
 & toto & 22.2 & 22.2 & 14.5 & 13.9 & 13.9 & 14.3 & 14.4 & \textbf{13.9}\textsuperscript{*} & \textbf{12.0}\textsuperscript{*} \\
 & chr$_2$ & 20.4 & 21.1 & 15.9 & 16.4 & 16.0 & 15.9 & 16.0 & 16.3 & \textbf{12.6}\textsuperscript{*} \\
 & moi$_2$ & 21.4 & 21.4 & 12.6 & 12.4 & 12.8 & 12.6 & 12.6 & \textbf{11.9}\textsuperscript{*} & \textbf{12.1}\textsuperscript{*} \\
\midrule
\multirow{6}{*}{rohlik} & tim & 36.1 & 30.3 & 33.4 & 26.8 & 26.4 & \textbf{26.1} & 26.1 & 26.7 & 26.6 \\
 & chr & 27.4 & 27.2 & 24.7 & 24.6 & 24.6 & 24.7 & 24.6 & \textbf{24.5}\textsuperscript{*} & \textbf{24.3}\textsuperscript{*} \\
 & moi & 36.2 & 31.2 & 33.5 & 33.6 & 33.4 & 32.9 & 32.6 & \textbf{32.5} & \textbf{28.4}\textsuperscript{*} \\
 & toto & 28.2 & 28.2 & 25.1 & 25.1 & 25.1 & 25.2 & 25.2 & \textbf{25.0}\textsuperscript{*} & \textbf{24.9}\textsuperscript{*} \\
 & chr$_2$ & 26.6 & 27.1 & 23.8 & 23.5 & 23.5 & 23.5 & 23.5 & \textbf{23.4}\textsuperscript{*} & \textbf{23.5}\textsuperscript{*} \\
 & moi$_2$ & 34.9 & 29.2 & 32.2 & 32.4 & 32.1 & 31.9 & 31.8 & \textbf{31.5} & \textbf{28.0}\textsuperscript{*} \\
\midrule
\multirow{6}{*}{bizitobs} & tim & 49.5 & 51.9 & 42.4 & 43.9 & 42.5 & 43.7 & 44.6 & 42.5 & \textbf{35.7}\textsuperscript{*} \\
 & chr & 24.6 & 25.0 & 25.7 & 28.2 & 25.7 & 25.7 & 25.7 & \textbf{24.6}\textsuperscript{*} & 26.7 \\
 & moi & 51.5 & 50.9 & 43.5 & 44.8 & 43.1 & 42.6 & 42.7 & 43.3 & \textbf{38.4}\textsuperscript{*} \\
 & toto & 39.0 & 39.0 & 40.9 & 44.0 & 39.0 & 41.6 & 41.0 & 40.4 & \textbf{36.6}\textsuperscript{*} \\
 & chr$_2$ & \textbf{23.6} & 25.3 & 31.0 & 23.6 & 24.4 & 24.4 & 24.4 & 25.9 & 25.8 \\
 & moi$_2$ & 27.1 & 27.1 & 27.7 & 27.6 & 27.9 & 28.4 & 28.2 & \textbf{26.7}\textsuperscript{*} & \textbf{26.6}\textsuperscript{*} \\
\midrule
\multirow{6}{*}{favorita} & tim & 24.9 & 23.2 & 21.8 & 21.3 & 21.7 & 21.6 & 21.5 & \textbf{21.1}\textsuperscript{*} & \textbf{21.0}\textsuperscript{*} \\
 & chr & 17.5 & \textbf{17.4} & 17.6 & 17.6 & 17.4 & 17.6 & 17.6 & 17.5 & 17.6 \\
 & moi & 20.6 & 20.6 & \textbf{19.8} & 19.8 & 20.0 & 19.8 & 19.8 & 19.9 & 19.9 \\
 & toto & 17.4 & 17.4 & 17.7 & 17.6 & 17.4 & 17.6 & 17.5 & 17.5 & \textbf{17.4} \\
 & chr$_2$ & 14.4 & 14.6 & 15.2 & 14.6 & 14.5 & 14.7 & 14.6 & \textbf{14.4}\textsuperscript{*} & 14.5 \\
 & moi$_2$ & 16.5 & 16.5 & 16.6 & 16.6 & 16.5 & 16.6 & 16.6 & 16.6 & \textbf{16.3}\textsuperscript{*} \\
\midrule
\multirow{6}{*}{m5} & tim & 52.5 & 51.4 & 51.0 & 51.1 & 51.0 & \textbf{50.9} & 50.9 & 51.1 & 52.4 \\
 & chr & 50.3 & 49.9 & 50.3 & 50.3 & 50.3 & 50.3 & 50.3 & \textbf{50.3} & 51.1 \\
 & moi & 52.2 & 52.2 & 52.1 & 52.1 & 52.1 & 52.0 & 52.0 & 52.2 & \textbf{52.0}\textsuperscript{*} \\
 & toto & 49.7 & 49.7 & 50.1 & 49.7 & 50.1 & 50.1 & 50.1 & \textbf{49.7} & 49.9 \\
 & chr$_2$ & 50.4 & 49.5 & 51.2 & 49.6 & 49.7 & 49.6 & 49.6 & \textbf{49.6} & \textbf{49.3}\textsuperscript{*} \\
 & moi$_2$ & \textbf{49.0} & 49.0 & 49.3 & 49.1 & 49.3 & 49.3 & 49.3 & 49.1 & 49.4 \\
\bottomrule
\end{tabular}}
\end{table*}

\end{document}